\documentclass[conference]{IEEEtran}
\usepackage{times}
\usepackage[table]{xcolor}

\usepackage[numbers]{natbib}
\usepackage[pagebackref=true,breaklinks=true,bookmarks=true,colorlinks]{hyperref}
\usepackage{url}
\usepackage{tikz} 
\usepackage{amsmath}
\usepackage{amssymb}
\usepackage{multirow}
\usepackage{multicol}
\usepackage{mathrsfs}
\usepackage{etoolbox}
\usepackage{xspace}
\usepackage{graphicx}
\usepackage{caption}
\usepackage{subcaption}
\usepackage{adjustbox}
\usepackage{xfrac}
\usepackage{listings}
\usepackage{float}
\usepackage{booktabs}
\usepackage{bm}
\usepackage{amsfonts}
\usepackage{amsthm}
\usepackage{mdframed}
\usepackage{algorithm}
\usepackage{algpseudocode}
\usepackage{multirow}
\usepackage{tabularx}
\usepackage{makecell}
\usepackage{pifont}
\usepackage{arydshln} 
\usepackage{etoc}
\usepackage{fvextra}
\usepackage{marvosym} 

\DeclareCaptionType{code}[Code][List of Codes]

\newcommand{\cmark}{\textcolor{green!70!black}{\ding{51}}} 
\newcommand{\xmark}{\textcolor{red!75!black}{\ding{55}}} 
\definecolor{touchblue}{HTML}{234AE1}
\definecolor{gtgreen}{HTML}{75FB4C}
\newcommand{\tb}[1]{\textcolor{touchblue}{#1}}
\newcommand{\kw}[1]{\textcolor{black!60}{#1}}
\newcommand{\kwd}[1]{\textcolor{black}{\textbf{#1}}}

\newtheorem{proposition}{Proposition}
\newtheorem{remark}{Remark}

\newcommand{\ie}{\textit{i}.\textit{e}., }
\newcommand{\eg}{\textit{e}.\textit{g}., }

\begin{document}

\title{TouchGuide: Inference-Time Steering of Visuomotor Policies via Touch Guidance}

\author{Zhemeng Zhang$^{1,2*}$\quad Jiahua Ma$^{3*}$\quad Xincheng Yang$^{2*}$\\ Xin Wen$^{3\dagger}$\quad Yuzhi Zhang$^{3\dagger}$\quad Boyan Li$^{1\dagger}$\quad Yiran Qin$^{4\ddagger}$\quad Jin Liu$^{1,2}$\quad Can Zhao$^{1}$\\ Li Kang$^{1,5}$\quad Haoqin Hong$^{6}$\quad Zhenfei Yin$^{4}$\quad Philip Torr$^{4}$\quad Hao Su$^{7}$\quad Ruimao Zhang$^{3}$\quad Daolin Ma$^{1,2\text{\Letter}}$ \vspace{1mm}\\

$^1$Shanghai Jiao Tong University\quad $^2$Xense Robotics\quad $^3$Sun Yat-sen University\quad $^4$Oxford\\ \,\,\,$^5$Shanghai AI Laboratory\quad $^6$University of Science and Technology of China\quad $^7$UCSD \vspace{1mm}\\

\,\,\,$^*$Equal contribution\quad $^\dagger$Equal contribution\quad $^\ddagger$Project leader\quad $^\text{\Letter}$Corresponding author \\
\,\,\,\texttt{martelzhang@gmail.com} \quad \texttt{\href{https://martelzhang.github.io/touchguide}{martelzhang.github.io/touchguide}}
}

\IEEEpeerreviewmaketitle
\twocolumn[{
    \renewcommand\twocolumn[1][]{#1}%
        \maketitle
        \vspace{-5mm}
	\begin{center}

\vspace{-0cm}
    \includegraphics[width=\textwidth]{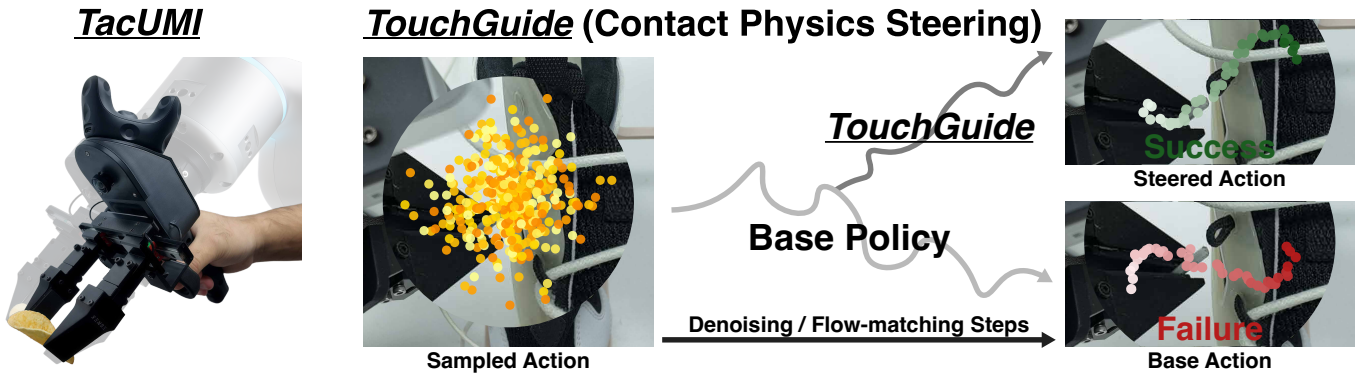}
    \captionof{figure}{\textbf{TacUMI} is a low-cost yet high-precision handheld data collection system that provides direct tactile feedback through a rigid mechanical coupling. \textbf{TouchGuide} is a multi-modal fusion paradigm that steers a visuomotor policy via touch guidance during denoising or flow matching, producing actions that better adhere to contact physics without retraining the base policy.}
    \label{fig:insight}
    \end{center}
}]

\begin{abstract}
Fine-grained and contact-rich manipulation remain challenging for robots, largely due to the underutilization of tactile feedback. To address this, we introduce TouchGuide, a novel cross-policy visuo-tactile fusion paradigm that fuses modalities within a low-dimensional action space. Specifically, TouchGuide operates in two stages to guide a pre-trained diffusion or flow-matching visuomotor policy at inference time. First, the policy produces a coarse, visually-plausible action using only visual inputs during early sampling. Second, a task-specific Contact Physical Model (CPM) provides touch guidance to steer and refine the action, ensuring it aligns with realistic physical contact conditions. Trained through contrastive learning on limited expert demonstrations, the CPM provides a tactile-informed feasibility score to steer the sampling process toward refined actions that satisfy physical contact constraints. Furthermore, to facilitate TouchGuide training with high-quality and cost-effective data, we introduce TacUMI, a data collection system. TacUMI achieves a favorable trade-off between precision and affordability; by leveraging rigid fingertips, it obtains direct tactile feedback, thereby enabling the collection of reliable tactile data. Extensive experiments on five challenging contact-rich tasks, such as shoe lacing and chip handover, show that TouchGuide consistently and significantly outperforms state-of-the-art visuo-tactile policies. Videos and other visualizations can be found on the project website: \texttt{\href{https://martelzhang.github.io/touchguide}{https://martelzhang.github.io/touchguide}}.
\end{abstract}

\section{Introduction}

Imitation learning~\cite{zhao2023act, ze2024dp3, chi2025diffusion, ma2025cdp, wang2025gaudp} has enabled robots to perform complex manipulation tasks~\cite{qin2025robofactory}, such as folding clothes and preparing coffee~\cite{intelligence2025pi06, hkummlab2025kai0, black2024pi0visionlanguageactionflowmodel, barbany2025bifold}. However, many intuitive fine-grained and contact-rich skills (\eg shoe lacing or handing over fragile chips) remain challenging. These tasks demand a policy that effectively integration of visual and tactile information: visual information provides global scene context and task understanding, while tactile feedback supplies localized, contact-specific information essential for fine-grained control. This motivates a central question: \emph{How should visual and tactile information be fused to improve policy planning in fine-grained and contact-rich manipulation?}

Existing visuo-tactile policy models generally follow one of two dominant fusion strategies: feature-level~\cite{xue2025reactive, wei2024ensuring, cheng2025omnivtla, bi2025vlatouch} or policy-level~\cite{chen2025policyconsensus, cao2025compose} fusion. While policy-level methods can mitigate the modality-dominance issue in feature-level concatenation caused by sparse features, they typically require training multiple single-modality policies, which makes it difficult to capture the cross-modal correlations that are critical for tasks demanding tight multimodal synergy.

To address this fundamental limitation, we introduce \textbf{TouchGuide}, a novel cross-policy visuo-tactile fusion paradigm. Departing from conventional computer vision approaches that focus on ``fusing multi-modal data per se", we rethink the problem from a robotic control perspective. Our \textbf{core idea} is to extract task-relevant semantics from each modality and perform fusion within the low-dimensional action space in a way that can benefits robotic manipulation. Concretely, 
TouchGuide steers a pre-trained visuomotor policy based on diffusion or flow matching (e.g., the action expert in $\pi_{0.5}$) during inference. The process begins by generating a visually feasible, coarse action from visual observation alone in the early sampling phase. Then, in the later phase, a task-specific Contact Physical Model (CPM) injects touch guidance, refining the action to better satisfy real-world physical contact constraints.
The CPM is trained via contrastive learning on limited expert demonstrations to directly output a feasibility score. This score assesses whether the action comply with the contact physical constraints implied by the current tactile observations in task-specific environment, thereby generating a more robust steering signal. Notably, CPM can be applied across multiple policies, including VLAs~\cite{intelligence2025pi_, black2024pi0visionlanguageactionflowmodel, shukor2025smolvla, wen2024diffusionvla}, diffusion policies~\cite{chi2025diffusion, ze2024dp3, wang2025gaudp, ma2025cdp}, and flow-matching policies~\cite{jiang2025streaming, zhang2025flowpolicy}.

Moreover, in fine-grained and contact-rich manipulation, policy execution performance depends heavily on the quality of expert demonstrations. Teleoperation, due to high latency and indirect haptic feedback, hinders intuitive control. While handheld devices such as UMI~\cite{chi2024umi} provide a platform-decoupled solution for data collection, existing systems relying on SLAM-based localization or motion capture (MoCap) cannot simultaneously achieve both high accuracy and low cost.
To this end, we introduce \textbf{TacUMI}, designed according to the following principles: (1) Cost-effective (base cost \textbf{\$720}, tactile sensors are additional, using Xense sensors) and lightweight (\textbf{540\,g}), while providing the high-precision data required for fine-grained manipulation. (2) Direct tactile feedback enabled by rigid fingertips. (3) A unified interface for synchronized acquisition of high-resolution visual and tactile signals.

Our main contributions are threefold:
\begin{itemize}
    \item We propose \textbf{TouchGuide}, a novel cross-policy fusion paradigm for fine-grained and contact-rich manipulation.
    \item We develop \textbf{TacUMI}, a cost-effective yet high-precision data collection system with direct tactile feedback.
    \item Experiments on five challenging fine-grained and contact-rich tasks show that our \textbf{TouchGuide} can outperform other SOTA visuo-tactile methods and generalize across policies and tactile modalities effectively.
\end{itemize}

\section{Related Work}

\begin{figure*}[!t]
  \centering
  \includegraphics[width=\textwidth]{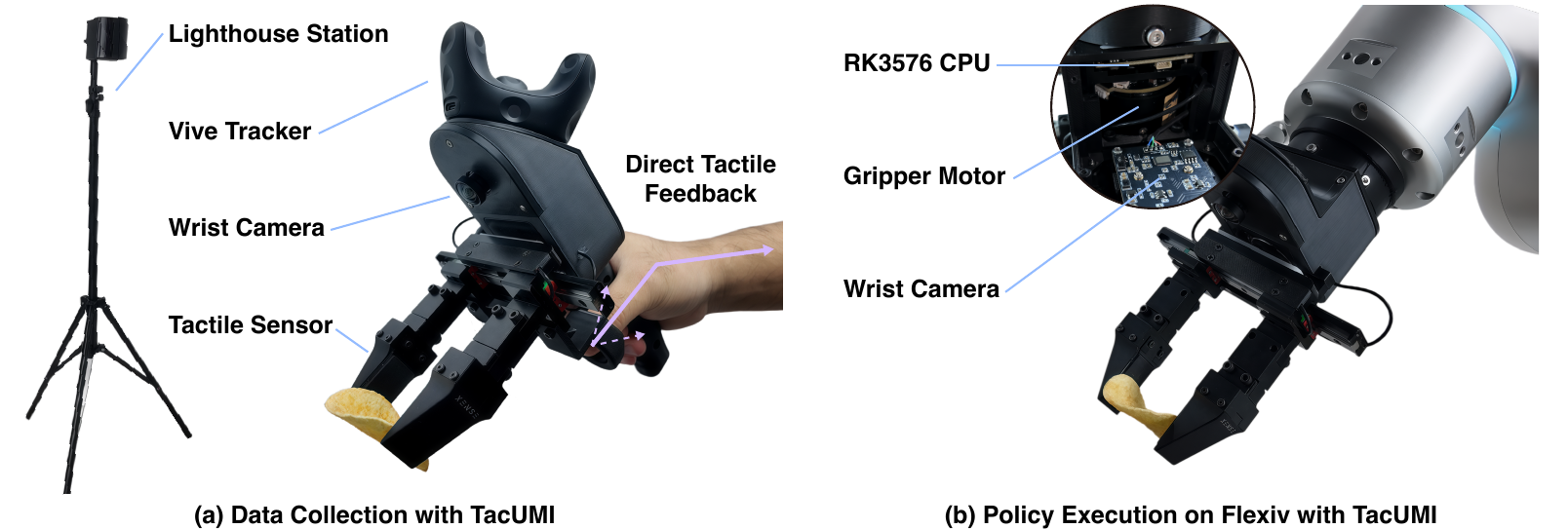}
  \caption{Overview of \textbf{TacUMI} data collection system. (a) TacUMI (Collection-side) uses a Vive tracker for localization to obtain accurate end-effector poses, while the operator receives direct tactile feedback. (b) During policy inference, we use an execution-side device that is structurally identical to the collection-side TacUMI, coupled to different robot arms via an adapter.
}
  \label{fig:tacumi}
  \vspace{-3mm}
\end{figure*}

\subsection{Hardware and Data Collection System}
\subsubsection{Tactile Sensors}
Optical tactile sensors~\cite{yuan2017gelsight, taylor2022gelslim, lambeta2020digit, ren2023mctac, lin20239dtact, lin2022dtact, kuppuswamy2020softbubble, xense} have advanced rapidly in recent years and have been widely adopted, owing to their ability to provide rich information (\ie force and torque can be indirectly inferred from tactile images~\cite{zhao2024ifem2, zhang2019effective}). Considering cost-effectiveness and ease of deployment, we selected the Xense tactile sensor~\cite{xense}. The sensor offers software force estimation interfaces~\cite{xensesdk}, making it possible to evaluate our system algorithms across multiple tactile modalities.

\subsubsection{Data Collection System}
Teleoperation is a common choice for data collection, from ALOHA-style leader–follower systems~\cite{zhao2023act, mu2024robotwin, aldaco2024aloha2, chen2025robotwin2, wu2024gello, fu2024mobilealoha, buamanee2024biact} to peripheral-driven control such as 3D mice~\cite{dhat20243dmice}, VR controllers~\cite{xue2025reactive, iyer2024openteach}, and hand tracking~\cite{qin2023anyteleop}. However, contact-rich and precise manipulation remains challenging due to missing tactile feedback; existing solutions mainly provide indirect haptic cues via VR visualization~\cite{xue2025reactive, kamijo2024learning, ding2025bunnyvisionpro}.

Some UMI-like~\cite{chi2024umi, zhaxizhuoma2025fastumi} handheld gripper data collection devices offer a data collection paradigm that operates independently of the physical robot~\cite{zhu2025touch, liu2025fastumi100k, zhaxizhuoma2025fastumi, xu2025dexumi}. This type of handheld data-collection gripper, which operates without the physical robot, can provide haptic feedback directly, and many tactile data collection systems use this paradigm~\cite{xu2025exumi, liu2025vitamin, zhu2025touch, wu2025freetacman, yu2025demooctopi15, helmut2025farm, li2025vitaminb, choi2026umift}. ViTaMIn~\cite{liu2025vitamin} and Touch in the Wild~\cite{zhu2025touch} adopt SLAM-based localization; however, the resulting accuracy is insufficient for many precise tasks. Some works (\eg exUMI~\cite{xu2025exumi}, ViTaMIn-B~\cite{li2025vitaminb}) attach a VR Motion Tracker (\eg VR controller of Meta Quest 3) to the gripper for localization, but this also introduces a notable drawback: it requires wearing a VR headset, which can impose a substantial burden during long-duration data collection. FreeTacMan~\cite{wu2025freetacman} and FARM~\cite{helmut2025farm} use motion-capture systems, achieving high accuracy while remaining lightweight. However, they impose stricter requirements on the capture environment, and the high cost of motion-capture equipment makes large-scale data collection difficult.

In contrast, our \textbf{TacUMI} uses Vive Trackers~\cite{vivetracker} for localization, reducing cost as much as possible while still meeting the accuracy requirements of precise manipulation tasks, thereby enabling large-scale data collection. Meanwhile, we minimize the linkage structure between the fingers and the gripper to obtain more direct haptic feedback, improving the quality of the tactile data.

\subsection{Visuo-tactile Policies}
Visuo-tactile policies are crucial for precise manipulation, and prior works~\cite{huang20243dvitac, xue2025reactive, zhang2025vtla, cheng2025omnivtla, hao2025tla, huang2025tactilevla, bi2025vlatouch, suresh2024neuralfeels, ding2023adaptivevisualtactile, lee2019makingsense, chen2022visuotactiletransformer, george2025vital, yang2022touchandgo, dave2024multimodalvisualtactile, chen2025policyconsensus} have continuously explored this direction. Most of these policies (\eg 3D-ViTac~\cite{huang20243dvitac}, VTLA~\cite{zhang2025vtla}, FreeTacMan~\cite{wu2025freetacman}) rely on simple feature-level concatenation, which can cause the policy to overemphasize a single modality~\cite{chen2025policyconsensus}. Policy-level composition methods such as PolicyConsensus~\cite{chen2025policyconsensus} and stage-wise fusion approaches like RDP~\cite{xue2025reactive} alleviate this issue to some extent, but they do not perform multi-modal fusion in the action space. In contrast, our method, \textbf{TouchGuide}, introduces Contact Physical Model (CPM) to efficiently fuse modalities in the action space, steering the policy distribution toward the real distribution (see Sec.~\ref{sec:method:steering}).

\subsection{Inference-time Steering}
Inspired by classifier guidance~\cite{dhariwal2021diffusion}, some recent works~\cite{du2025dynaguide, sun2025lpb, wang2025inferencetimesteering, zhang2025inferencetimescaling} have begun to adopt inference-time steering for robotic manipulation, which is an effective way to guide actions toward a target distribution during inference. DynaGuide~\cite{du2025dynaguide} and LPB~\cite{sun2025lpb} guide visuomotor policies via a dynamics model; however, this paradigm is ineffective for fine-grained manipulation, because future observations can become entirely different due to small errors in the action. In contrast, TouchGuide focuses on the current contact physics and can be integrated with a wide range of visuomotor policies (\eg DP~\cite{chi2025diffusion} and $\pi_{0.5}$~\cite{intelligence2025pi_}), yielding clear improvements over dynamics-based approaches (results are presented in Sec.~\ref{sec:exp}).

\section{Data Collection System: TacUMI}
\label{sec:tacumi}
\subsection{System Overview}
As shown in Fig.~\ref{fig:tacumi}, TacUMI is a UMI-like~\cite{chi2024umi} handheld gripper that collects tactile data without relying on proprioception. Its development follows three key characteristics:

\textbf{Accuracy and Low-cost.} TacUMI is localized using a Vive tracker with two Lighthouse base stations, providing reliable and accurate pose estimates. In contrast, UMI~\cite{chi2024umi} relies on SLAM-based tracking; for fine-grained tactile manipulation tasks which requires high precision, its accuracy is insufficient and tracking can be easily lost in feature-sparse scenes. Notably, a setup with one Vive tracker and two Lighthouse base stations has a base cost of only about \textbf{\$720}, maintaining low cost while providing accurate localization. A detailed comparison of data collection quality across different methods is provided in the experiments (see Sec.~\ref{sec:exp:system}).

\textbf{Direct Tactile Feedback.} Given that collecting tactile data for fine-grained manipulation critically depends on real-time tactile feedback, teleoperation presents a fundamental challenge by introducing a delay that impedes the operator's ability to perceive and respond to touch signals promptly. Although some teleoperation methods, such as TactAR~\cite{xue2025reactive}, provide visualized tactile feedback, this relies on human learning. After becoming accustomed to the interface, operators often ignore the visual tactile cues and instead act based on experience~\cite{xue2025reactive}. In contrast, this gripper-finger design enables direct tactile feedback, facilitating the collection of high-quality data.  

\textbf{Usability and Synchronization.} TacUMI adopts a unified hardware-software interface. With a single aviation cable, it streams all observations in real time, including visual and tactile signals and proprioceptive states. Our data collection is compatible with the LeRobot~\cite{cadene2024lerobot} format and uses one software pipeline to acquire all modalities in a unified timestamped stream, enabling effective software-level synchronization. TacUMI captures visual $\mathbf{V}_t \in \mathbb{R}^{640\times480\times3}$ and tactile images $\mathbf{T}_t^\text{image} \in \mathbb{R}^{200\times350\times3}$ at 30Hz, sufficient for most fine-grained manipulation tasks. We also provide force observations $\mathbf{T}_t^\text{force} \in \mathbb{R}^{20\times35\times3}$ as a dense 3D force field; in our algorithm, it is handled the same as tactile images, except for the input dimensionality. An onboard RK3576 CPU runs the XenseSDK~\cite{xensesdk} to compute this force field online from the tactile images, without requiring external processing, as data are being captured. Since tactile sensing is modality-specific (\ie tactile images are not shared across sensors for training), this capability is crucial for scalable tactile data collection.

More details on hardware selection, design choices, and system parameters are provided in the Appendix~\ref{sec:appen:hardware}.

\subsection{Data Processing for Training}
We define $\{W\}$ as the world frame provided by Vive-tracker Lighthouse, $\{T\}$ as the Vive-tracker frame, $\{E\}$ as the end-effector (ee) frame and $\{B\}$ as the robot base frame. Before data collection, we align the robot base frame with the world frame, \ie 
the transformation matrix $^WT_B \in SE(3)$ should be obtained, and $^TT_E$ can be computed from the 3D model file. The data sequence collected by the Vive tracker after the coordinate transformation is $\{^BT_E^{(0)} \dots ^BT_E^{(i)}\}$, \ie the end-effector pose in the robot base frame. Similar to FastUMI~\cite{zhaxizhuoma2025fastumi}, the data sequence can be expressed as $\{\mathbf{p}^{(i)}_{\text{ee}}, \mathbf{R}^{(i)}_{\text{ee}}\}$, where $\mathbf{p}^{(i)}_{\text{ee}}$ and $\mathbf{R}^{(i)}_{\text{ee}}$ are the end-effector position and orientation in the robot base frame, respectively.

For policy training, we use relative ee pose $\{\mathbf{p}_{\text{rel}}^{(i)}, \mathbf{R}_{\text{rel}}^{(i)}\}$ as the dataset format, which is considered more effective than using absolute ee pose~\cite{chi2024umi}. This can be expressed as follows:
\begin{align}
    \mathbf{p}_{\text{rel}}^{(i)} &= {\mathbf{R}_{\text{ee}}^{(0)\top}} (\mathbf{p}_{\text{ee}}^{(i)} - \mathbf{p}_{\text{ee}}^{(0)}), \\
    \mathbf{R}_{\text{rel}}^{(i)} &= {\mathbf{R}_{\text{ee}}^{(0)\top}} \mathbf{R}_{\text{ee}}^{(i)},
\end{align}
where $\{\mathbf{p}_{\text{ee}}^{(0)}, \mathbf{R}_{\text{ee}}^{(0)}\}$ denotes the last observation state in an action chunk. Unlike UMI~\cite{chi2024umi}, we use the absolute ee pose $\{\mathbf{p}^{(i)}_{\text{ee}}, \mathbf{R}^{(i)}_{\text{ee}}\}$ as the robot state, with the goal of enabling the policy to perceive its pose in the real workspace. Moreover, since our model is designed to provide inference-time steering for both DP~\cite{chi2025diffusion} and $\pi_{0.5}$~\cite{intelligence2025pi_}, and $\pi_{0.5}$ typically conditions on a single observation step, using relative state would make the state always the identity matrix, which would render proprioception ineffective.

\section{Method: TouchGuide}

\begin{figure*}[!th]
  \centering
  \includegraphics[width=\textwidth]{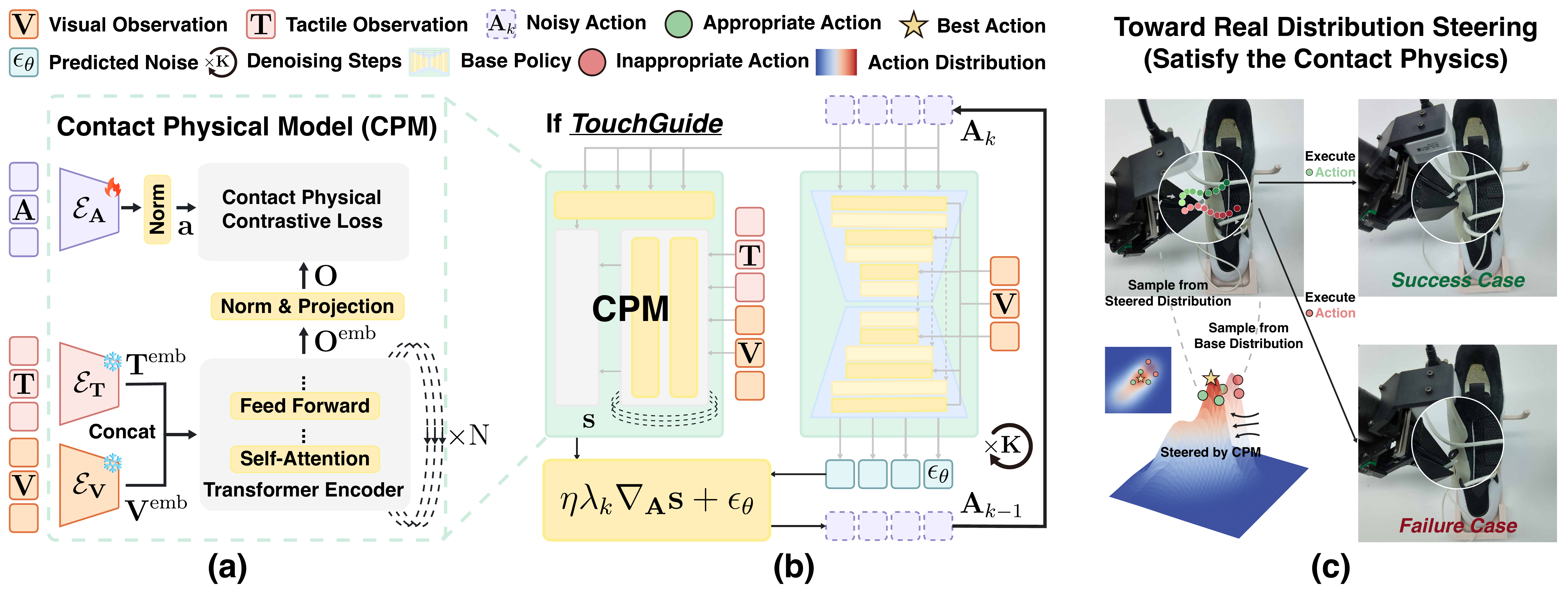}
  \caption{Overview of \textbf{TouchGuide} framework. (a) The architecture of the task-specific Contact Physical Model (CPM). (b) During inference, the CPM serves as an external model that steers the base policy’s action generation within the sampling process using a feasibility score. (c) In action space, TouchGuide can be viewed as a form of contact-physics steering that steers the policy distribution toward the real distribution.
}
  \label{fig:cpm}
  \vspace{-3mm}
\end{figure*}

In this section, we introduce TouchGuide, a two-stage cross-policy multimodal fusion paradigm based on classifier guidance~\cite{dhariwal2021diffusion} (Sec.~\ref{sec:method:pcg}), inspired by recent advances in inference-time steering~\cite{sun2025lpb, du2025dynaguide, wang2025inferencetimesteering}.
TouchGuide performs inference-time steering on a pre-trained diffusion-based or flow-matching-based (\eg action expert of $\pi_{0.5}$) visuomotor policy $\pi_\theta(\mathbf{A}_t \,|\, \mathbf{V}_t)$. During the early phase of the denoising or flow-matching process, the policy generates an initial coarse action based solely on visual observation, ensuring visual feasibility without yet enforcing tactile constraints. In the later phase, a task-specific Contact Physical Model (CPM, detailed in Sec.~\ref{sec:method:cpmtraining}) provides touch guidance via a feasibility score, which steers the action toward better satisfaction of real-world physical contact.
The complete inference-time guidance process of TouchGuide is detailed in Sec.~\ref{sec:method:steering}.

\subsection{Preliminaries: Classifier Guidance}
\label{sec:method:pcg}
\textbf{Diffusion Model.}
Following the notation in prior work~\cite{dhariwal2021diffusion}, classifier guidance can be written as follows:
\begin{align}
    \hat{\epsilon}_\theta(x_t) = \epsilon_\theta(x_t) - \eta \sqrt{1 - \bar{\alpha}_t} \nabla_{x_t} \log p_\phi(y | x_t).
    \label{eq:diff_cg}
\end{align}

\textbf{Flow Matching.}
Since prior work has not explicitly derived classifier guidance for flow matching~\cite{zheng2023guidedfm, feng2025guidancefm}, we provide a brief introduction to the corresponding flow-matching formulation of classifier guidance.
\begin{proposition}
\label{prop:cg4fm}
The classifier guidance for flow matching can be formulated as follows:
\begin{align}
    \hat{u}_\theta(x_t)
    = u_\theta(x_t)
    - \eta \frac{t}{1 - t} \nabla_{x_t} \log p_\phi(y | x_t),
    \label{eq:fm_cg}
\end{align}
where $u_\theta(x_t)$ is the unconditional velocity field, $\phi$ parameterizes the auxiliary noise-dependent classifier, $y$ is the target class label (\eg additional condition), and $\eta$ is the guidance scale. The term $t / (1 - t)$ acts as a time-dependent weighting coefficient derived from the specific flow definition. See the detailed deviration in Appendix~\ref{sec:appen:cg4fm}.
\end{proposition}
\begin{remark}
\label{remark:vlas}
For VLAs, we can achieve tactile modality fusion by steering their action expert. Specifically, since the action expert of $\pi_{0.5}$ is flow-matching based, we can apply Eq.~\ref{eq:fm_cg} to steer the action sampling process (see Eq.~\ref{eq:fm_steering} in Sec.~\ref{sec:method:steering}).
\end{remark}

\subsection{Contact Physical Model}
\label{sec:method:cpmtraining}

At time step $t$, Contact Physical Model (CPM, see Fig.~\ref{fig:cpm}(a)) outputs a task-specific feasibility score conditioned on the tactile feedback $\mathbf{T}_t$, the visual observation $\mathbf{V}_t$  and the coarse action $\mathbf{A}_t$ generated by the early phase of pre-trained policy $\pi_\theta(\mathbf{A}_t \,|\, \mathbf{V}_t)$. Specifically, we use pretrained DINOv2 as both the tactile and visual encoder to obtain the  tactile and visual embedings, \ie $\mathbf{T}_t^{\text{emb}}$ and $\mathbf{V}^{\text{emb}}_{t}$ respectively, as it provides effective representations of robotic observations~\cite{du2025dynaguide, zhou2024dino} (see Appendix~\ref{sec:appen:obsenc} for encoder comparison experiments). Then, we fuse these two embedings using N-layer Transformer encoder $\mathtt{TranEnc}(\cdot)_{\times \text{N}}$ and finally, yield the latent observation $\mathbf{O}_t$:
\begin{align}
    \mathbf{O}_t^{\text{emb}} = \mathtt{TranEnc}(\mathtt{concat}[\mathbf{T}_t^{\text{emb}}, \mathbf{V}_t^{\text{emb}}])_{\times \text{N}} \xrightarrow[]{\text{L2 Norm}} \mathbf{O}_t.
\end{align}
Similarly, we can obtain the latent action $\mathbf{a}_t$ by:
\begin{align}
    \mathbf{A}_t^{\text{emb}} = \mathcal{E}_{\mathbf{A}}(\mathbf{A}_t) \xrightarrow[]{\text{L2 Norm}} \mathbf{a}_t,
\end{align}
where $\mathcal{E}_{\mathbf{A}}$ is the action encoder (1D CNN + MLP) and $\mathbf{A}_t^{\text{emb}}$ is the action embedding.
The feasibility score $\mathbf{s}$ represents the cosine similarity between the observation and the action in the latent space~\cite{chen2020simclr}, and is defined as follows:
\begin{align}
    \mathbf{s} = s_\phi(\mathbf{V}_t, \mathbf{T}_t, \mathbf{A}_t) = \mathbf{O}_t^\top \mathbf{a}_t.
    \label{eq:feasibility_score_definition}
\end{align}
We provide a derivation of our design for the feasibility score $\mathbf{s}$, along with its physical interpretation (see Appendix~\ref{sec:appen:fs_derivation}).

\textbf{Training Strategy.} We use the limited expert demonstrations to train our task-specific CPM. Given the visual-tactile observation (\ie $\mathbf{V}_t$ and $\mathbf{T}_t$) and an ground-truth action $\mathbf{A}_t$, we can yield the latent observation $\mathbf{O}_t$ and latent action $\mathbf{a}_t$. Subsequently, we establish the positive samples using latent observation-action pairs from the same time step $\{(\mathbf{O}_t, \mathbf{a}_t)\}$, and negative samples using pairs from different time steps $\{(\mathbf{O}_{t + \Delta t}, \mathbf{a}_t), (\mathbf{O}_t, \mathbf{a}_{t + \Delta t})\}$ . Finally, a standard contrastive learning loss $\mathcal{L}_{\text{CPM}}$ is adopted:
\begin{equation}
    \mathcal{L}_{\text{CPM}}(\mathbf{O}_t, \mathbf{a}_t) = \frac{1}{2}(\mathcal{L}_{\mathbf{O} \to \mathbf{a}} + \mathcal{L}_{\mathbf{a} \to \mathbf{O}}),
\end{equation}
\vspace{-4mm}
\begin{align}
    \mathcal{L}_{\mathbf{O} \to \mathbf{a}} &= -\frac{1}{M}\sum_{i = 1}^{M} \log \frac{\exp(\mathbf{O}_i^\top \mathbf{a}_i / \tau)}{\sum_{j = 1}^{M}\exp(\mathbf{O}_i^\top\mathbf{a}_j / \tau)},  \\
    \mathcal{L}_{\mathbf{a} \to \mathbf{O}} &= -\frac{1}{M}\sum_{i = 1}^{M} \log \frac{\exp(\mathbf{a}_i^\top \mathbf{O}_i / \tau)}{\sum_{j = 1}^{M}\exp(\mathbf{a}_i^\top\mathbf{O}_j / \tau)},
\end{align}
where $\tau$ is a learnable temperature parameter and $M$ denotes the number of samples used during training.
In order to enable the CPM to effectively operate in a noisy action space, which is generated by the early phase of the pretrained policy during inference, we enhance the ground-truth action $\mathbf{A}_t$ by adding noise following a geometric distribution using the same noise scheduler as DP~\cite{du2025dynaguide}; for flow-matching policies, we instead apply linear interpolation.

\begin{algorithm}[!ht]
\caption{TouchGuide (Inference Time)}
\label{alg:touchguide}
\begin{algorithmic}[1]
\Require Base policy $\pi_\theta$, Contact Physical Model $s_\phi$, total steps $K$, guidance steps $K_{\text{TouchGuide}}$, guidance scale $\eta$, visual observation $\mathbf{V}_t$, tactile observation $\mathbf{T}_t$
\State $\mathbf{A}^K_t \leftarrow$ Sample from $\mathcal{N}(0, I)$
\For{$k = K, \dots, 1$}
    \If{$\pi_\theta \in$ Diffusion-based} \Comment{\eg DP, DP3}
        \State $\epsilon \leftarrow \pi_\theta(\mathbf{V})$
        \If{$k \leq K_{\text{TouchGuide}}$}
            \State $\mathbf{s} \leftarrow s_\phi(\mathbf{V}_t, \mathbf{T}_t, \mathbf{A}^k_t)$
            \State $\hat{\epsilon} \leftarrow \epsilon - \eta \sqrt{1 - \bar{\alpha}_k} \nabla_{\mathbf{A}^k_t}\mathbf{s}$
            \State $\mathbf{A}^{k - 1}_t \leftarrow$ Update $\mathbf{A}^{k}_t$ using $\hat{\epsilon}$
        \Else
            \State $\mathbf{A}_t^{k - 1} \leftarrow$ Update $\mathbf{A}_t^{k}$ using ${\epsilon}$
        \EndIf
    \ElsIf{$\pi_\theta \in$ Flow-matching} \Comment{\eg $\pi_{0.5}$}
        \State $k \leftarrow$ Scale $[K, 0]$ to $[1, 0]$
        \State $u \leftarrow \pi_\theta(\mathbf{V}_t)$
        \If{$k \leq K_{\text{TouchGuide}}$}
            \State $\mathbf{s} \leftarrow s_\phi(\mathbf{V}_t, \mathbf{T}_t, \mathbf{A}_t^k)$
            \State $\hat{u} \leftarrow u - \eta \frac{k}{1 - k} \nabla_{\mathbf{A}_t^k}\mathbf{s}$ 
            \State $\mathbf{A}_t^{k - 1} \leftarrow$ Update $\mathbf{A}_t^{k}$ using $\hat{u}$
        \Else
            \State $\mathbf{A}_t^{k - 1} \leftarrow$ Update $\mathbf{A}_t^{k}$ using ${u}$
        \EndIf
    \EndIf
\EndFor
\State \Return $\mathbf{A}_t^0$
\end{algorithmic}
\end{algorithm}
\vspace{-3mm}

\subsection{Steering the Policies via Touch Guidance}

\label{sec:method:steering}
\begin{figure}[!ht]
  \centering
  \includegraphics[width=\linewidth]{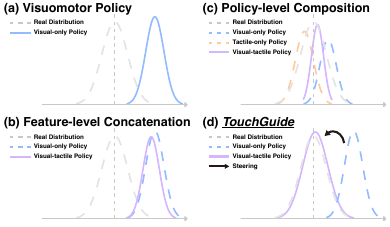}
  \caption{Comparison of policy distributions in action space.}
  \label{fig:distribution}
  \vspace{-4mm}
\end{figure}

Fig.~\ref{fig:cpm}(b) illustrates the entire TouchGuide pipeline, where touch guidance is used to steer the sampling process of base policy to generate actions that better adhere to the contact physics. Following~\cite{du2025dynaguide, sun2025lpb}, both diffusion models and flow-matching models can be guided by the gradients  $\nabla_{x_t}\log p_\phi(y | x_t)$ provided by a classifier toward a particular set of samples that maximizes $p_\phi(y | x_t)$ (see Eq.~\ref{eq:diff_cg},~\ref{eq:fm_cg})~\cite{dhariwal2021diffusion, zheng2023guidedfm, feng2025guidancefm}. 
In our case, $\log p_\phi(y | x_t)$ corresponds to the feasibility score $\mathbf{s}$ produced by the CPM. With this formulation, touch guidance for flow-matching (\eg action expert of $\pi_{0.5}$) and diffusion policies can be respectively expressed by the following two equations:
\begin{equation}
\hat{\epsilon}_\theta = \epsilon_\theta(\mathbf{A}_t^k, \mathbf{V}_t) - \eta \sqrt{1 - \bar{\alpha}_k}\nabla_{\mathbf{A}_t^k} s_\phi(\mathbf{V}_t, \mathbf{T}_t, \mathbf{A}_t^k),
\label{eq:dp_steering}
\end{equation}
\begin{equation}
    \hat{u}_\theta = u_\theta(\mathbf{A}_t^k, \mathbf{V}_t) - \eta \frac{k}{1 - k} \nabla_{\mathbf{A}_t^k} s_\phi(\mathbf{V}_t, \mathbf{T}_t, \mathbf{A}_t^k),
    \label{eq:fm_steering}
\end{equation}
where $\eta$ is the guidance scale, $\epsilon_\theta(\cdot)$ and $u_\theta(\cdot)$ are the noise predictor in diffusion policy and velocity predictor in flow-matching policy, respectively. The detailed \textbf{TouchGuide} procedure is provided in Alg.~\ref{alg:touchguide}.

As shown in Fig.~\ref{fig:cpm}(c),
our TouchGuide can be viewed as using the CPM to steer the policy distribution $\mathcal{Q}_{\text{policy}}$ toward the real distribution $\mathcal{Q}_{\text{real}}$, which refers to the optimal distribution to accomplish the task in the real world.
Therefore, sampling actions from the CPM-steered $\mathcal{Q}_\text{policy}$ is more likely to produce actions that better adhere to the contact physics.
In Fig.~\ref{fig:distribution}, we illustrate the policy distributions of TouchGuide and previous methods (including feature-level concatenation~\cite{xue2025reactive, wei2024ensuring, cheng2025omnivtla, bi2025vlatouch} and policy-level composition~\cite{cao2025compose, chen2025policyconsensus}) in the action space. 
For feature-level concatenation, although tactile sensing provides additional information, the model may still overly attend to the visual modality (since tactile signals often exhibit limited variation in many cases), and thus its policy distribution $\mathcal{Q}_{\text{policy}}$ cannot be explicitly driven toward the real distribution $\mathcal{Q}_{\text{real}}$. Policy-level composition, though an improvement over feature-level methods, still falls short in enabling effective multi-modal fusion within the action space $\mathcal{A}$. Its distribution operates multiplicatively in $\mathcal{A}$, and it lacks an explicit mechanism to steer the $\mathcal{Q}_{\text{policy}}$ toward $\mathcal{Q}_{\text{real}}$. To address these limitations, we introduce a CPM. Our method assesses the physical feasibility of both tactile observations and noisy actions, thereby providing explicit guidance for steering in $\mathcal{A}$. Further physical interpretations of how CPM steers the action distribution are provided in the Appendix~\ref{sec:appen:fs_derivation}.

\section{Experiments}
\label{sec:exp}
\begin{figure*}[!ht]
  \centering
  \includegraphics[width=\textwidth]{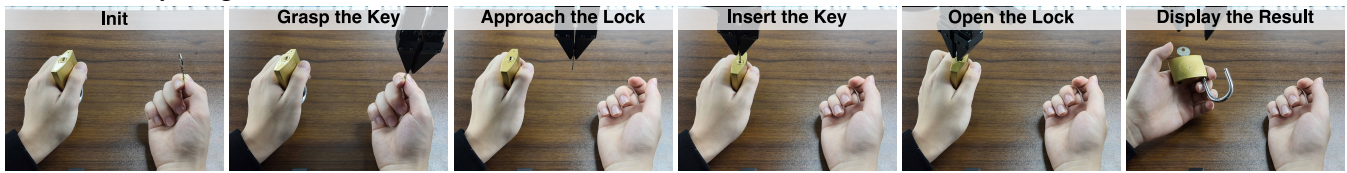}
  \caption{Five experiment tasks including Shoe Lacing, Chip Handover, Cucumber Peeling, Vase Wiping, and Lock Opening.}
  \label{fig:task}
  \vspace{-3mm}
\end{figure*}

In this section, we evaluate the proposed TouchGuide framework and the TacUMI data-collection system across five challenging tasks. Specifically, our experiments are designed to answer the following questions: 
\textbf{Q1.} How does tactile feedback improve fine-grained robotic manipulation?
\textbf{Q2.} How does TouchGuide perform compared to prior visuo-tactile policies?
\textbf{Q3.} Does TouchGuide generalize across different robots, tactile sensors, and policies?
\textbf{Q4.} Is the TouchGuide fusion framework robust and generalizable under visual domain shifts and occlusions?
\textbf{Q5.} Can TacUMI efficiently provide high-quality data for precise contact-rich tasks?

\begin{table*}[!ht]
\centering
\caption{Policy performance comparison on five challenging fine-grained, contact-rich tasks.
}
\label{tab:results}
\renewcommand{\arraystretch}{1.3}
\newcolumntype{Y}{>{\centering\arraybackslash}X} 
\begin{tabularx}{\textwidth}{@{} ll *{6}{Y} @{}}
\toprule
 & & \multicolumn{3}{c}{Bi-Arx5 (Dual-arm)} & \multicolumn{2}{c}{Flexiv Rizon4 (Single-arm)} & \\ 
\cmidrule(lr){3-5} \cmidrule(lr){6-7}
 &  & \makecell{Shoe Lacing \\ (100 Demos)} & \makecell{Chip Handover \\ (50 Demos)} & \makecell{Cucum. Peeling \\ (50 Demos)} & \makecell{Vase Wiping \\ (30 Demos)} & \makecell{Lock Opening \\ (20 Demos)} & Average \\ \midrule
\multirow{8}{*}{DP} & Diffusion Policy & 0\% & 5\% & 0.500 & 0.265 & 0\% & 16.3\% \\
 & DP w/ Tactile Observation & 0\% & 15\% & 0.520 & 0.240 & 5\% & 19.2\% \\
 & SafeDiff & 0\% & 10\% & 0.635 & 0.325 & 5\% & 22.2\% \\
 & RDP & 0\% & 20\% & 0.740 & 0.475 & 10\% & 30.3\% \\
 & Policy Consensus & 0\% & 15\% & 0.630 & 0.405 & 5\% & 24.7\% \\
 & Tactile Dynamics & 0\% & 15\% & 0.695 & 0.335 & 0\% & 23.6\% \\
 & \textbf{TouchGuide (Force)} & 0\% & \textbf{30\%} & 0.805 & 0.510 & 15\% & 35.3\% \\ 
 & \textbf{TouchGuide (Tactile Img.)}& 0\% & 25\% & \textbf{0.810} & \textbf{0.550} & \textbf{20\%} & \textbf{36.2\%} \\ \midrule
\multirow{5}{*}{$\pi_{0.5}$} & $\pi_{0.5}$ & 20\% & 25\% & 0.785 & 0.360 & 20\% & 35.9\% \\
 & $\pi_{0.5}$ w/ Tactile Observation & 20\% & 15\% & 0.760 & 0.305 & 10\% & 30.3\% \\
 & Tactile Dynamics & 10\% & 30\% & 0.815 & 0.395 & 20\% & 36.2\% \\
 & \textbf{TouchGuide (Force)} & 25\% & 40\% & 0.955 & 0.590 & \textbf{30\%} & 49.9\% \\ 
 & \textbf{TouchGuide (Tactile Img.)} & \textbf{35\%} & \textbf{60\%} & \textbf{0.975} & \textbf{0.675} & \textbf{30\%} & \textbf{58.0\%} \\ \bottomrule
\end{tabularx}
\vspace{-3mm}
\end{table*}

\subsection{Experimental Setup}
\label{sec:exp:setup}
\subsubsection{Hardware}
For dual-arm tasks, we employ the Bi-ARX5~\cite{arx5}, while the Flexiv Rizon4~\cite{flexiv} integrated with TacUMI is used for single-arm tasks. Given the high symmetry of the gripper's tactile sensing, we attach tactile sensors only to the outer finger of each arm during dual-arm experiments. This design reduces dataset size and enhances training efficiency.

\subsubsection{Tasks}
As shown in Fig.~\ref{fig:task}, we design five challenging contact-rich tasks. To thoroughly evaluate the performance of our paradigm, these tasks include long-horizon, precise manipulation (\eg Shoe Lacing), collaborative handover of fragile objects (\eg Chip Handover), and low-data (only 20 demos), high-precision manipulation (\eg Lock Opening). More detailed task descriptions and examples of the evaluation metrics are provided in Appendix~\ref{sec:appen:tasks}.

\textbf{Shoe Lacing.} The robot needs to grasp the shoelace and sequentially thread it through four eyelets (two on each side). The task is considered successful if the robot threads the shoelace through all four eyelets within the allotted time.

\textbf{Chip Handover.} The robot needs to grasp a potato chip, execute a collaborative in-air handover, and finally place it into a designated container. The task is considered successful if, within the allotted time, the chip is placed in the container without any breaking.

\textbf{Cucumber Peeling.} The robot must grasp both a peeler and a cucumber, after which the two robotic arms collaborate to peel the cucumber. Task score is assessed based on the ratio of the length of peel removed by the robot to that removed by a human expert. The resulting score is quantized into one of seven discrete values: 0, 0.1, 0.3, 0.5, 0.7, 0.9, or 1.

\textbf{Vase Wiping.} The robot needs to pick up a blackboard eraser from the table, approach the vase held by a human, and erase the marker traces on its surface. The task score is defined similarly to Cucumber Peeling.

\textbf{Lock Opening.} The robot executes a sequential manipulation task involving a human partner: grasping a key held by the human, inserting it into a lock also held by the human, and finally turning the key to open the lock. The task is considered successful if the robot maintains hold of the key without dropping it and opens the lock within the allotted time.

\subsubsection{Baselines}
We adopt two state-of-the-art (SOTA) visuomotor policies (\ie DP~\cite{chi2025diffusion} and $\pi_{0.5}$~\cite{intelligence2025pi_}) as the base policy for TouchGuide, and use several visuo-tactile policies (\eg RDP~\cite{xue2025reactive}, Policy Consensus~\cite{chen2025policyconsensus} and SafeDiff~\cite{wei2024ensuring}) as the baselines for comparison. Specifically, we include Tactile Dynamics as one of our baselines. This method is derived from Latent Policy Barrier~\cite{sun2025lpb} and employs a dynamics model to jointly predict tactile and visual observations. Details on the touch guidance setup of Tactile Dynamics for $\pi_{0.5}$ can be found in Sec.~\ref{sec:method:pcg}. For TouchGuide, we use two tactile modalities (\ie force and tactile image) to thoroughly validate the cross-modality capability of our paradigm. The specific baseline configurations are detailed in  Appendix~\ref{sec:appen:baselines}.
For these five tasks, each baseline is evaluated over 20 trials with moderate randomization of the initial environment conditions, and the results are reported in Table~\ref{tab:results}.

\subsection{Main Results and Analysis}
\textbf{Tactile sensing provides rich and accurate in-hand state information, which is crucial for fine-grained manipulation (Q1).} Tactile sensing not only provides rich force information but also yields accurate in-hand state information. For fine-grained manipulation tasks such as shoe lacing, chip handover, and lock opening, this in-hand state information is particularly critical, as it determines the end-effector pose required for threading through an eyelet or executing a reliable handover. In contrast, such in-hand states are difficult to infer accurately from vision alone and are often only weakly observable through specular highlights. 
Table~\ref{tab:results} shows that TouchGuide yields clear performance gains over DP and $\pi_{0.5}$ (\textbf{16.3\% $\rightarrow$ 36.2\%, 35.9\% $\rightarrow$ 58.0\%}). This highlights the importance of accurate in-hand state estimation for precise manipulation.

\textbf{TouchGuide explicitly steers the base policy in action space via touch guidance, yielding performance gains over current SOTA visuo-tactile policies (Q2).}
Existing methods mainly target rapid reactions to tactile changes (\eg RDP~\cite{xue2025reactive} and PolicyConsensus~\cite{chen2025policyconsensus}); however, RDP relies on low-dim tactile embeddings that discard in-hand state cues, hindering fine-grained manipulation.
In many tasks, initial contact is decisive: for Lock Opening and Shoe Lacing, small errors at first contact can drop the key or severely misalign it, pushing the system into out-of-distribution states that are especially problematic for imitation learning.
In contrast, our method uses a CPM to assess the feasibility of noisy actions proposed by the base policy and steers the policy at inference time using the resulting feasibility scores.
Evaluating physical consistency between actions and tactile signals is crucial for producing appropriate contacts.
Table~\ref{tab:results} supports this claim: TouchGuide (\textbf{36.2\%}) improves over other SOTA visuo-tactile policies, including RDP~\cite{xue2025reactive} (30.3\%) and PolicyConsensus~\cite{chen2025policyconsensus} (24.7\%). Meanwhile, TouchGuide adds negligible inference time overhead (see Appendix~\ref{sec:appen:tg_imple:performace}).

\begin{figure*}[!t]
  \centering
  \includegraphics[width=\textwidth]{figure/tsne.pdf}
  \vspace{-6mm}
  \caption{Action distribution visualization and evaluation setups on Chip Handover. (a1,a2) t-SNE of $\mathbf{a}_\text{base}$, $\mathbf{a}_\text{real}$, $\mathbf{a}_\text{steered}$. (b1) Training / Normal setup. (b2) Hard setup with larger displacement and rotation.}
  \label{fig:tsne}
  \vspace{-3mm}
\end{figure*}

\textbf{TouchGuide demonstrates strong generalization, and can be applied across different robots, tactile modalities, and policies (Q3).} Unlike prior visuo-tactile policies, TouchGuide augments a visuomotor policy with touch guidance in a fully decoupled manner, requiring no retraining to the base policy. 
As a result, it can inherit the strong capability of the base policy. 
For example, TouchGuide built on $\pi_{0.5}$ achieves solid performance on a high-precision task such as Lock Opening with only \textbf{20 demonstrations}.
Moreover, TouchGuide achieves strong performance across different robots (\eg Bi-ARX5 and Flexiv Rizon4), different policies (\eg DP and $\pi_{0.5}$), and different tactile modalities (\eg force and tactile image). Detailed results are reported in Table~\ref{tab:results}. More baseline failure cases are provided in Appendix~\ref{sec:appen:failurecase}.

\subsection{Ablation Studies}
\label{sec:exp:ablation}
\subsubsection{Noise Pretraining Ablation} Since the task-specific CPM takes noisy actions as input at inference time, we add noise into the action using the same noise scheduler for diffusion models, or via linear interpolation for flow-matching models. The number of noising steps is sampled from a geometric distribution. As shown in Table~\ref{tab:noise_pretrain}, this key setting can effectively improve the task success rate for TouchGuide (\textbf{39.17\% $\rightarrow$ 62.50\%}).

\begin{table}[!ht]
\vspace{-1mm}
\centering
\caption{Ablation study on noise pretraining.}
\label{tab:noise_pretrain}
\resizebox{\linewidth}{!}{
\begin{tabular}{lcccc}
\toprule
& Chip & Cucumber & Lock & \\
& Handover & Peeling & Opening & Average \\
\midrule
Ours w/o Noise Pretraining & 30\% & 0.725 & 15\% & 39.17\% \\
\textbf{Ours w/ Noise Pretraining} & \textbf{60\%} & \textbf{0.975} & \textbf{30\%} & \textbf{62.50\%} \\
\bottomrule
\end{tabular}
}
\vspace{-3mm}
\end{table}

\subsubsection{CPM Modality Ablation} For most fine-grained manipulation tasks, tactile and visual signals are sparse at different time periods. For example, in the Cucumber Peeling task, tactile feedback is sparse when the robot holds the peeler and moves toward the cucumber. During peeling, however, small motion variations lead to pronounced changes in tactile feedback, while visual information becomes relatively sparse. Therefore, we argue that for task-specific CPM, the visual and tactile modalities are equally important
(see Table~\ref{tab:tg_modality_ablation}). The results show without vision, the success rate drops from \textbf{62.50\%} to \textbf{43.33\%}, and without touch, it decreases to \textbf{43.50\%}.

\begin{table}[!ht]
\vspace{-1mm}
\centering
\caption{Ablation study on CPM input modalities.}
\label{tab:tg_modality_ablation}
\resizebox{\linewidth}{!}{
\begin{tabular}{lcccc}
\toprule
& Chip & Cucumber & Lock & \\
& Handover & Peeling & Opening & Average \\
\midrule
Ours w/o Vision & 30\% & 0.850 & 15\% & 43.33\% \\
Ours w/o Touch  & 35\% & 0.755 & 20\% & 43.50\% \\
\textbf{Ours w/ Vision \& Touch}    & \textbf{60\%} & \textbf{0.975} & \textbf{30\%} & \textbf{62.50\%} \\
\bottomrule
\end{tabular}
}
\vspace{-3mm}
\end{table}

Also,we provide ablation studies on the guidance scale $\eta$ and guidance steps $K_\text{TouchGuide}$ in Alg.~\ref{alg:touchguide} (see Appendix~\ref{sec:appen:steerhyper}).

\begin{figure*}[!t]
  \centering
  \includegraphics[width=\textwidth]{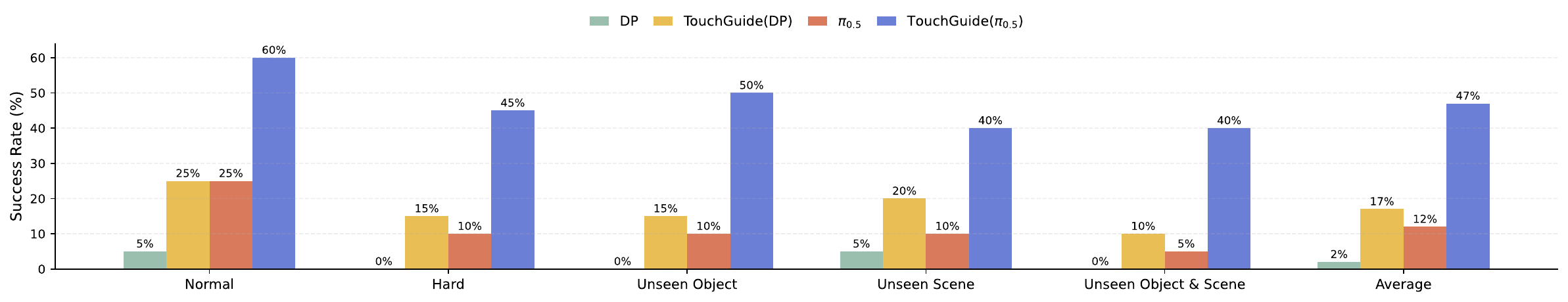}
  \vspace{-4mm}
  \caption{Generalization evaluation on Chip Handover across Normal, Hard, Unseen Object, Unseen Scene, and Unseen Object \& Scene conditions. TouchGuide consistently and substantially improves the base policy under all settings.}
  \label{fig:bar_chart}
  \vspace{-4mm}
\end{figure*}

\subsection{Generalization and Robustness}
\label{sec:exp:generalization}
\begin{figure}[!t]
  \centering
  \includegraphics[width=\linewidth]{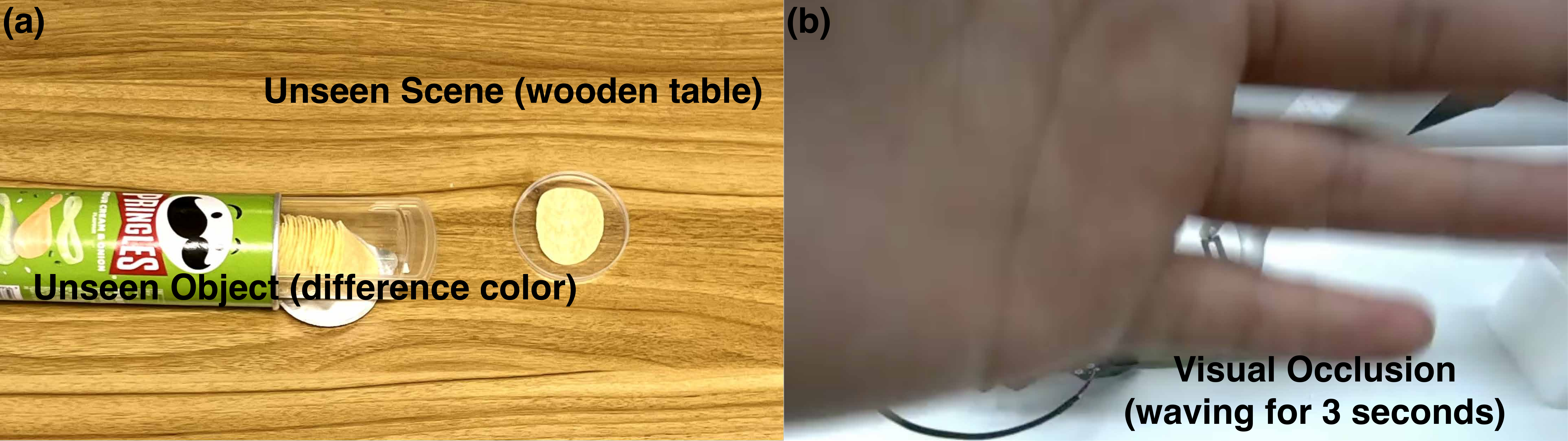}
  \vspace{-6mm}
  \caption{Additional experimental settings. (a) Unseen Object and Unseen Scene setups. (b) Visual Occlusion setup.}
  \label{fig:unseen}
  \vspace{-3mm}
\end{figure}

\textbf{TouchGuide is robust under visual domain shifts and occlusions (Q4).}
As shown in Fig.~\ref{fig:bar_chart}, TouchGuide consistently improves the base policy across unseen objects, unseen scenes, and their combination (Fig.~\ref{fig:unseen}(a)), indicating that touch guidance compensates for visual distribution shifts by providing complementary contact information. Table~\ref{tab:visual_occlusion} further demonstrates robustness to transient visual occlusion (Fig.~\ref{fig:unseen}(b)). Fig.~\ref{fig:tsne}(a1,a2) confirms that $\mathbf{a}_\text{steered}$ aligns more closely with $\mathbf{a}_\text{real}$ than $\mathbf{a}_\text{base}$ (Additional details are provided in Appendix~\ref{sec:appen:tsne}).

\begin{table}[!t]
\centering
\caption{Robustness evaluation under visual occlusion at different contact phases in the Cucumber Peeling task.}
\label{tab:visual_occlusion}
\resizebox{\linewidth}{!}{
\begin{tabular}{lccccc}
\toprule
 & None & Before & During & After & Average \\
\midrule
TouchGuide($\pi_{0.5}$) & 0.975 & 0.955 & 0.910 & 0.960 & 0.950 \\
\bottomrule
\end{tabular}
}
\vspace{-3mm}
\end{table}

\subsection{System Comparison and User Study}
\label{sec:exp:system}
\begin{table}[!t]
\centering
\caption{Success rate comparison of data collection systems on the Lock Opening task.}
\label{tab:dc_succ}
\renewcommand{\arraystretch}{1.15}
\setlength{\tabcolsep}{10pt}
\begin{tabular}{lccc}
\toprule
Method & \textbf{TacUMI} & UMI (SLAM-based) & VR Teleop \\
\midrule
$\pi_{0.5}$           & 20\% & 0\% & 5\% \\
\textbf{TouchGuide}   & \textbf{30\%} & 5\% & 15\% \\
\bottomrule
\end{tabular}
\vspace{-5mm}
\end{table}
\textbf{TacUMI provides accurate and reliable data that is well suited for robot learning (Q5).} We find that although VR teleoperation provides highly precise action (since it is read directly from the robot arm), teleoperating tasks, such as Lock Opening, are challenging for human demonstrators. Hesitation around the keyhole often degrades the data quality, making it unsuitable for robot learning. This issue is particularly pronounced in the low-data regime (20 demos), where such low-quality demonstrations can even hinder learning. In addition, SLAM-based UMI~\cite{chi2024umi} struggles to meet the requirements of fine-grained manipulation tasks due to the limited accuracy of its SLAM module. In order to evaluate the effectiveness of our TacUMI, we trained policies using the data collected by all involved data collection systems and evaluated them on the Lock Opening task. Table~\ref{tab:dc_succ} indicates that our TacUMI outperforms the other two data collection systems.

To evaluate the usability of our TacUMI, we recruited five volunteers working in robotics (two had prior experience collecting demonstrations using a handheld gripper, while three had no such experience). Each volunteer collected 20 valid demonstrations using VR Teleop and UMI with three different tracking methods (\eg SLAM-based, VR Motion Tracker, and \textbf{TacUMI}) on Lock Opening task. We used the following metrics: \textbf{Attempts} (the number of trials required to obtain the target number of demonstrations), \textbf{Length} (the dataset duration excluding preparation and post-processing time, which helps assess system usability), \textbf{Valid Rate} (Valid Data / Attempts), and \textbf{Satisfaction} (each volunteer’s subjective usability rating of the system).
As shown in Table~\ref{tab:dc_user}, although Teleop achieves a high valid rate, the tactile feedback is not directly available, which leads to excessively long demonstrations and may degrade policy performance. SLAM-based tracking yields a very low valid rate. The VR Motion Tracker requires wearing an VR headset, making it unsuitable for prolonged data collection, and occlusions can easily cause tracking loss if the user is not constantly attentive during collection. \textbf{TacUMI} demonstrates excellent usability and receives high satisfaction ratings from the volunteers.
More comparisons with existing data collection systems are provided in Appendix~\ref{sec:appen:hardware:comparison}.

\begin{table}[!ht]
\centering
\caption{User study on data collection systems.}
\label{tab:dc_user}
\resizebox{\linewidth}{!}{
\begin{tabular}{lcccc}
\toprule
Method & Attempts $\downarrow$ & Length $\downarrow$ & Valid Rate $\uparrow$ & Satisfaction $\uparrow$ \\
\midrule
VR Teleop         & \textbf{100} & 23.0\,min & \textbf{100.00\%} & 7.8 \\
SLAM-based        & 143 & 15.3\,min & 69.93\%  & 7.2 \\
VR Motion Tracker & 104 & 14.7\,min & 96.15\%  & 9.0 \\
\textbf{TacUMI (Ours)}     & \textbf{100} & \textbf{13.5\,min} & \textbf{100.00\%} & \textbf{9.6} \\
\bottomrule
\end{tabular}
}
\vspace{-3mm}
\end{table}

\section{Limitations and Future Work}
Although our TouchGuide paradigm enables tactile fusion without retraining the base policy, the CPM still requires task-specific training, which inevitably increases training time and cost. An important direction for future work is to learn more effective tactile representations and develop a non-task-specific CPM that can generalize across tasks without retraining.

\section{Conclusion}
In this paper, we propose a novel two-stage cross-policy multi-modal fusion paradigm (\ie \textbf{TouchGuide}), and an efficient data collection system (\ie \textbf{TacUMI}). TouchGuide steers the base policy via touch guidance to generate actions that better adhere to the contact physics, demonstrating strong performance on five challenging fine-grained manipulation tasks. TacUMI strikes an trade-off between cost and precision, minimizing overhead while maintaining high accuracy; moreover, its direct tactile feedback improves usability and helps operators collect high-quality tactile data more efficiently.

\section*{Acknowledgments}
This work was supported in part by the National Natural Science Foundation of China (NSFC) under Grants No. 12522202, No. 12272220 and No. 124B1038, and in part by Xense Robotics.

\bibliographystyle{plainnat}
\bibliography{references}

\clearpage
\newpage
\onecolumn
\appendix

\begingroup
  \etocsetnexttocdepth{subsubsection}
  \etocsettocstyle{\vspace{6pt}}{\vspace{10pt}}
  \renewcommand{\baselinestretch}{1.2}\selectfont
  \setlength{\parskip}{4pt}
  \localtableofcontents
\endgroup

\clearpage

\subsection{Classifier Guidance for Flow Matching (Proof of Proposition~\ref{prop:cg4fm})}
\label{sec:appen:cg4fm}
\textbf{TouchGuide} steers the action expert in VLAs. In particular, for $\pi_{0.5}$, the action expert is flow-matching-based, and Equation~\ref{eq:fm_steering} is adopted to perform contact physics steering. Eq.~\ref{eq:fm_steering} follows from Proposition~\ref{prop:cg4fm} and is used during policy action sampling (\ie at inference time). Prior work~\cite{zheng2023guidedfm, feng2025guidancefm} has primarily focused on classifier-free guidance for flow matching, and detailed derivations of classifier guidance for flow matching are difficult to find. We therefore provide a brief derivation here to support the use of Eq.~\ref{eq:fm_steering} for inference-time steering in flow-matching-based models (\ie action expert of $\pi_{0.5}$). We next detail the derivation of Prop.~\ref{prop:cg4fm}.
\begin{proof}
Flow Matching relies on defining a probability path $p_t(x)$ that transforms a source distribution (\textit{e.g.}, noise) to a target data distribution $q(x_0)$. We define the conditional distribution $p_t(x|x_0)$ as a Gaussian path:
\begin{align}
    p_t(x|x_0) = \mathcal{N}(x|\alpha_t x_0,\sigma_t^2 I).
\end{align}
This implies that a sample $x$ at time $t$ can be expressed via the reparameterization:
\begin{align}
    x &= \alpha_t x_0 + \sigma_t \epsilon,
    \label{eq:reparam}
\end{align}
where $\epsilon \sim \mathcal{N}(0, I)$. We define the conditional vector field $u_t(x|x_0)$ as the time derivative of the sample trajectory. Differentiating Eq.~\ref{eq:reparam} with respect to time $t$, the following equation is obtained.
\begin{align}
    u_t(x|x_0) = \frac{\sigma_t^\prime}{\sigma_t}x + \left( \alpha_t^\prime - \frac{\sigma_t^\prime \alpha_t}{\sigma_t} \right)x_0.
    \label{eq:uxx1}
\end{align}

For the Gaussian distribution $p_t(x|x_0)$, the gradient of its log-density is given by
\begin{align}
    \nabla_x \log p_t(x|x_0) = -\frac{x - \alpha_t x_0}{\sigma_t^2}.
\end{align}
By substituting the above expression into Eq.~\ref{eq:uxx1}, the following formula is obtained
\begin{align}
    u_t(x|x_0) &= a_t x + b_t \nabla \log p_t(x | x_0),
\end{align}
where $a_t = \alpha_t^\prime / \alpha_t, b_t = (\alpha_t^\prime / \alpha_t - \sigma_t^\prime / \sigma_t) \sigma_t^2$. Thus,
\begin{align}
     u_t(x) &= \int u_t(x|x_0) \frac{p_t(x|x_0)q(x_0)}{p_t(x)} dx_0 \notag \\
     &= a_tx \int \frac{p_t(x|x_0)q(x_0)}{p_t(x)} dx_0 + b_t \int \nabla \log p_t(x | x_0) \frac{p_t(x|x_0)q(x_0)}{p_t(x)} dx_0 \notag \\
     &= a_t x \frac{\int p_t(x|x_0)q(x_0) dx_0}{p_t(x)} + b_t \int \frac{\nabla p_t (x | x_0)}{p_t(x|x_0)} \frac{p_t(x|x_0)q(x_0)}{p_t(x)} dx_0 \notag \\
     &= a_tx + b_t \frac{\nabla \int p_t(x|x_0)q(x_0)dx_0}{p_t(x)} \notag \\
     &= a_t x + b_t \nabla \log p_t(x).
\end{align}
For our flow matching policy $u_\theta(\cdot)$, since $u_\theta(\cdot)$ is trained to approximate $u_t(\cdot)$, we can obtain the following approximation:
\begin{align}
    u_\theta(x_t) \approx a_t x_t + b_t\nabla_{x_t} \log p_\theta(x_t).
\end{align}
By applying Bayes' theorem to the score function, we decompose the conditional score into a prior score and a likelihood score. It is straightforward to show that:
\begin{align}
    \nabla_{x_t}\log \hat{p}_\theta(x_t | y) &= \nabla_{x_t} \log p_\theta(x_t) + \nabla_{x_t} \log p_\phi(y | x_t) \notag \\
    &= \frac{1}{b_t} u_\theta(x_t) - \frac{a_t}{b_t} x_t + \nabla_{x_t} \log p_\phi(y|x_t).
\end{align}

Then, we can define velocity prediction with classifier guidance $\hat{u}_\theta$ which corresponds to the score of the joint distribution:
\begin{align}
    \hat{u}_\theta(x_t) := u_\theta(x_t) + b_t \nabla_{x_t} \log p_\phi(y|x_t).
\end{align}
We construct the conditional flow using the optimal transport path. Specifically, the intermediate state $x_t$ is obtained via linear interpolation with coefficients $\alpha_t = 1 - t, \sigma_t = t$. Empirically, we observe results consistent with classifier guidance~\cite{dhariwal2021diffusion}, finding that introducing a gradient scale $\eta$ significantly improves performance. Consequently, we arrive at the conclusion stated in Proposition~\ref{prop:cg4fm}.
\end{proof}

\subsection{Steering Hyperparameter Investigation}
\label{sec:appen:steerhyper}
In this section, we detail the selection of steering hyperparameter, including the guidance scale and guidance steps, as a supplement to the discussion in Sec.~\ref{sec:exp:ablation}.
\subsubsection{Ablation Study on Steering Hyperparameter}
\begin{figure*}[!ht]
  \centering
  \includegraphics[width=0.75\textwidth]{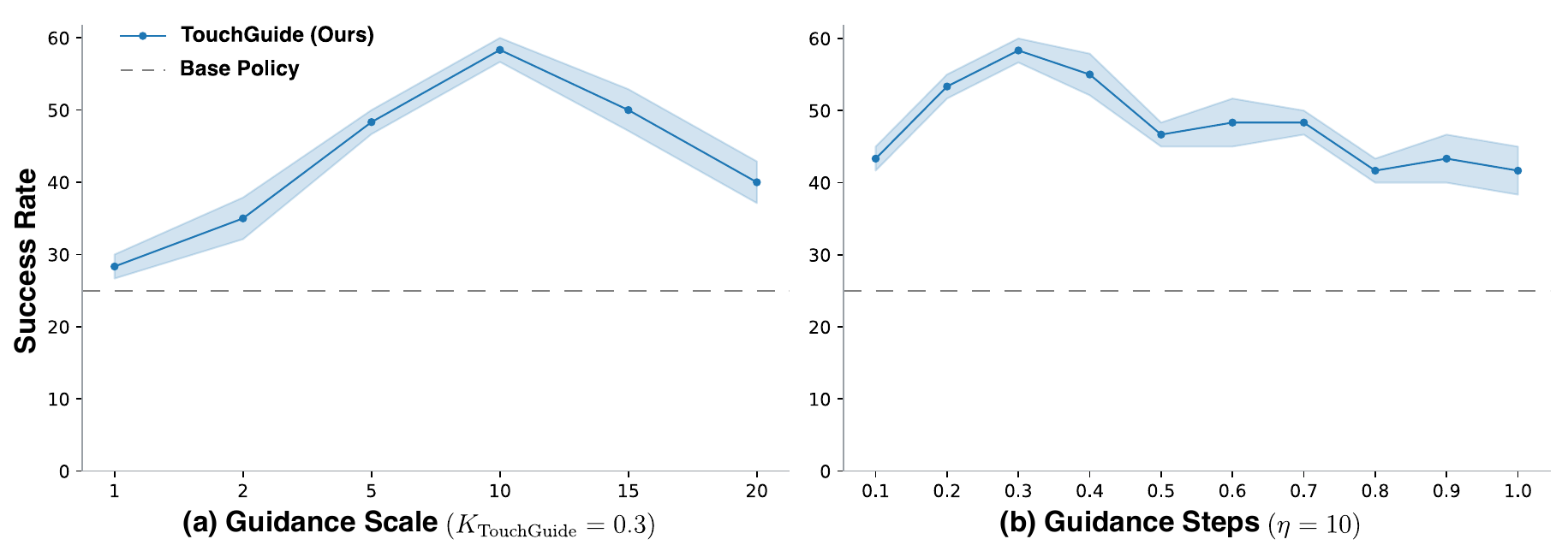}
  \caption{\textbf{Steering hyperparameter choice on Chip Handover task using $\pi_{0.5}$ as the base policy.} (a) Performance differences across guidance scale (with guidance step $K_\text{TouchGuide} = 0.3$). (b) Performance differences across guidance steps (with guidance scale $\eta = 10$).}
  \label{fig:hyperinvestigation}
  \vspace{-3mm}
\end{figure*}

To select TouchGuide steering hyperparameters, we conducted extensive experiments on the Chip Handover task using $\pi_{0.5}$~\cite{intelligence2025pi_} as the base policy. We primarily varied two hyperparameters (\ie guidance scale $\eta$ and guidance steps $K_\text{TouchGuide}$, for the detailed hyperparameter implementation, see Alg.~\ref{alg:touchguide}) by sweeping one while holding the other fixed. For each hyperparameter setting, we performed 3 repeats, each consisting 20 trials, and recorded the success rate. The results are shown in Fig.~\ref{fig:hyperinvestigation}, where the error bars indicate the standard error of the mean (SEM). The SEM $\sigma$ is computed as follows:
\begin{align}
    \bar{x} = \frac{1}{n}\sum_{i = 1}^{n} x_i, \quad s = \sqrt{\frac{1}{n - 1}\sum_{i = 1}^{n}(x_i - \bar{x})^2}, \quad \sigma = \frac{s}{\sqrt{n}}.
\end{align}

\textbf{Guidance Scale.} As shown in Fig.~\ref{fig:hyperinvestigation}(a), with $K_{\text{TouchGuide}} = 0.3$, the mean success rate reaches its maximum at a guidance scale $\eta$ of 10. If the guidance scale is too small, the contact physics constraint provides insufficient steering for the base policy. Conversely, an overly large guidance scale can cause overshooting, substantially perturbing the base policy’s original action sampling process.

\textbf{Guidance Steps.} Since we use $\pi_{0.5}$ as the base policy, its action expert is a flow-matching-based model whose action sampling steps are scaled to $[0, 1]$ with a step size of $0.1$. For example, $K_\text{TouchGuide} = 0.3$ means that contact physics steering is applied only during the last three flow-matching sampling steps. As shown in Fig.~\ref{fig:hyperinvestigation}(b), with a guidance scale of $\eta = 10$, the mean success rate peaks at $K_\text{TouchGuide} = 0.3$. This is because too few guidance steps do not give TouchGuide enough opportunities to steer the base policy, whereas too many steps introduce excessive noise into the actions fed to the CPM, which can in turn lead to erroneous guidance. Notably, for $\pi_{0.5}$, TouchGuide improves the base policy across all choices of guidance steps, highlighting the robustness of our TouchGuide visuo-tactile fusion paradigm.

Overall, as shown in Fig.~\ref{fig:hyperinvestigation}, TouchGuide is relatively robust to the choice of guidance scale and guidance steps, consistently improving the base policy’s performance across a wide range of settings.

\subsubsection{Steering Hyperparameter Used for Each Task.} As shown in Table~\ref{tab:steerhyper}, we report all steering hyperparameters used for the five challenging fine-grained, contact-rich tasks. For Diffusion Policy, we use a \texttt{DDPMScheduler} with 100 denoising steps. For $\pi_{0.5}$, we use linear interpolation with 10 flow-matching steps. Note that the steering hyperparameters should be tailored to the specific task, policy, and action representation. We caution against directly reusing the same large hyperparameter values. In practice, it is advisable to start from the smallest values and gradually increase them at the beginning of each experiment.

\begin{table}[!ht]
  \centering
  \caption{Hyperparameters for TouchGuide across tasks.}
  \label{tab:steerhyper}
  \renewcommand{\arraystretch}{1}
  \setlength{\tabcolsep}{8pt}
  \begin{tabular}{ccccccc}
    \toprule
     &  & \makecell{Shoe \\ Lacing} & \makecell{Chip \\ Handover} & \makecell{Cucumber \\ Peeling} & \makecell{Vase \\ Wiping} & \makecell{Lock \\ Opening} \\
    \midrule
    \multirow{2}{*}{Diffusion Policy} & $K_\text{TouchGuide}$ & 15  & 20  & 20  & 10  & 10 \\
    & $\eta$ & 3   & 4   & 4   & 4   & 4  \\
    \midrule
    \multirow{2}{*}{$\pi_{0.5}$}      & $K_\text{TouchGuide}$ & 0.2 & 0.3 & 0.3 & 0.3 & 0.3 \\
    & $\eta$ & 10  & 10  & 10  & 10  & 10 \\
    \bottomrule
  \end{tabular}
  \vspace{-3mm}
\end{table}

\subsection{Ablation Study on Observation Encoder}
\label{sec:appen:obsenc}

In this section, we analyze how replacing the visual and tactile encoders with alternative backbones affects the performance of the CPM. We evaluate each configuration in terms of the number of trainable parameters and task success rates. Notably, we focus on comparing DINOv2 with pretrained weights against ResNet18 co-trained from random initialization, when used as the visual and tactile encoders.

\begin{table*}[!ht]
\vspace{-1mm}
\centering
\small
\caption{Comparison of CPM performance under different visual and tactile encoders.}
\label{tab:encoder_ablation}
\begin{tabular}{lccccc}
\toprule
{Visual / Tactile Encoder} &
\makecell{Trainable \\ Parameters (CPM)} &
\makecell{Chip \\ Handover} &
\makecell{Cucumber \\ Peeling} &
\makecell{Lock \\ Opening} &
Average \\
\midrule
ResNet18 / ResNet18          & 141\,M & 50\% & 0.820 & 15\% & 49.0\% \\
DINOv2 / ResNet18            & 130\,M & \textbf{60\%} & 0.965 & 25\% & 60.5\% \\
\textbf{DINOv2 / DINOv2 (Ours)}       & \textbf{118\,M} & \textbf{60\%} & \textbf{0.975} & \textbf{30\%} & \textbf{62.5\%} \\
\bottomrule
\end{tabular}
\vspace{-3mm}
\end{table*}

As shown in Table~\ref{tab:encoder_ablation}, using the same pretrained DINOv2 \textbf{(frozen)} for both the visual and tactile encoders yields the highest success rate (\textbf{62.5\%}) while requiring the fewest trainable parameters (\textbf{118\,M}). In contrast, co-training two randomly initialized ResNet18 encoders results in the lowest success rate (49.0\%), which is somewhat surprising. We conjecture that, in our low-data regime, ResNet18 struggles to learn effective feature extractors from scratch. Moreover, the CPM primarily relies on effectively fusing the visual and tactile embeddings, a capability largely provided by the stacked Transformer encoders. Using a pretrained DINOv2 \textbf{(frozen)} visual encoder with a co-trained ResNet18 tactile encoder performs slightly worse than our setting (62.5\% $\rightarrow$ 60.5\%) while increasing the number of trainable parameters (130\,M). Therefore, we adopt pretrained DINOv2 \textbf{(frozen)} for both the visual and tactile encoders.

\subsection{Baseline Failure Cases Analysis}
\label{sec:appen:failurecase}

\begin{figure*}[!ht]
\vspace{-3mm}
  \centering
  \includegraphics[width=\textwidth]{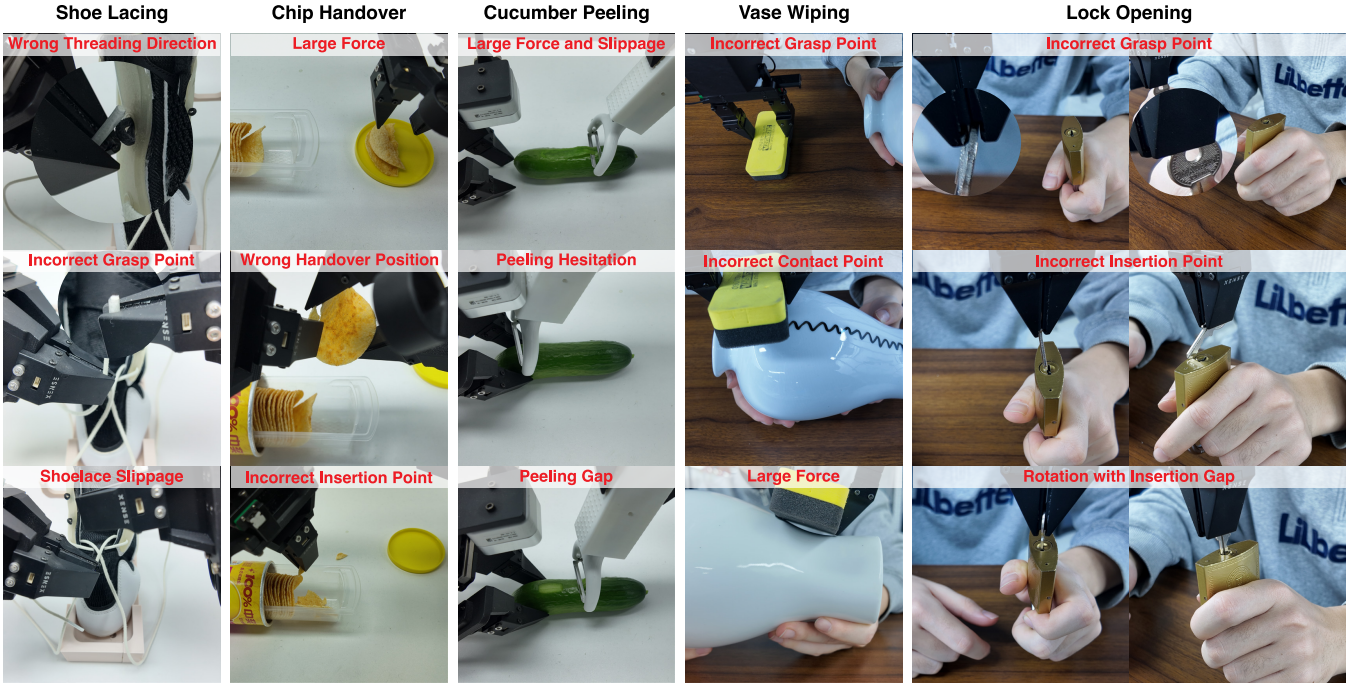}
  \caption{Common baseline failure cases for Shoe Lacing, Chip Handover, Cucumber Peeling, Vase Wiping, and Lock Opening.}
  \label{fig:baseline_failure}
  \vspace{-3mm}
\end{figure*}

In this section, we provide a detailed analysis of common baseline failure cases across five tasks, namely Shoe Lacing, Chip Handover, Cucumber Peeling, Vase Wiping, and Lock Opening. Fig.~\ref{fig:baseline_failure} summarizes representative failure cases of the baselines. Table~\ref{tab:results} reports the success rates of all baselines, Fig.~\ref{fig:task} illustrates the workflows of the five tasks, and Appendix~\ref{sec:appen:baselines} provides implementation details for each baseline.

\subsubsection{Shoe Lacing}
For the Shoe Lacing task, common failure cases include wrong threading direction, incorrect grasp point, and shoelace slippage.
These failures are largely caused by missing in-hand state information. 

\textbf{Wrong Threading Direction.} When threading the shoelace through an eyelet, the insertion angle often depends on the angle at which the gripper grasps the lace. This grasp angle is difficult to observe from both the wrist and head cameras. The wrist camera provides a top-down view, where the grasp angle is typically inferred from subtle specular highlights on the lace, making it highly sensitive to lighting conditions and not robust to illumination changes. For the head camera, the low image resolution and limited data often prevent the model from learning such subtle variations.

\textbf{Incorrect Grasp Point.} For incorrect grasp points, the Shoe Lacing task requires repositioning the shoelace to an appropriate grasp location before threading. Only the tip of the lace is rigid; if the gripper grasps too far back and picks up the flexible section, the lace cannot be inserted. Conversely, grasping too close to the tip may leave insufficient length for successful threading. In this setting, in-hand state information is crucial, as it can precisely identify where the lace is grasped and enable incremental adjustments toward a suitable grasp point.

\textbf{Shoelace Slippage.} Shoelace slippage is also a major source of failure. Without tactile feedback, it is difficult to assess whether the lace is securely grasped, especially when the two arms coordinate to reposition it. In the absence of in-hand state information, one gripper may release before the other has fully secured the shoelace, causing it to slip and leading to task failure.

As shown in Table~\ref{tab:results}, all DP-based policies achieve a success rate of \textbf{0}. This is likely because these models are relatively small, and with limited expert demonstrations they struggle to fit long-horizon, fine-grained manipulation behaviors. This also highlights an advantage of \textbf{TouchGuide}, namely its ability to integrate with different base policies in a cross-policy manner.

For $\pi_{0.5}$, we hypothesize that incorporating tactile input is difficult because the VLM backbone has not been pretrained with this modality. Consequently, with limited fine-tuning (\textbf{40k steps}), the model may not effectively learn the new modality, and the feature-level concatenation variant of $\pi_{0.5}$ does not outperform the vision-only $\pi_{0.5}$.

\subsubsection{Chip Handover} For the Chip Handover task, the primary failure cases include large force, wrong handover position, and incorrect insertion point. Wrong handover position and incorrect insertion point are mainly caused by missing in-hand state information, whereas large force primarily stems from the lack of tactile force feedback.

\textbf{Large Force.} Without force feedback, manipulating fragile objects becomes difficult, and even simple pick-and-place manipulation can be challenging. In the chip grasping setting, vision can only infer the applied force from the chip’s deformation, yet even subtle differences in deformation can lead to drastically different task outcomes.

\textbf{Wrong Handover Position.} For the Chip Handover task, the handover stage is arguably the most challenging. During handover, in-hand state can inform the appropriate handover orientation. Due to the chip’s irregular shape, only the relatively flat region near the edge is suitable for transfer. Adjusting the handover angle based on the chip’s grasp orientation is therefore critical for successful completion. However, this grasp orientation is difficult to infer from visual observations alone, whereas tactile sensing provides rich and accurate in-hand state that can effectively compensate for this limitation.

\textbf{Incorrect Insertion Point.} Similarly, tactile sensing can supplement in-hand state information and help the robot insert the chip into the chip canister at a more favorable orientation.

In our experiments (see Tab.~\ref{tab:results}), RDP~\cite{xue2025reactive} performs poorly. We attribute this to its use of PCA to compress tactile images into a 15-dimensional tactile embedding, which discards substantial in-hand state information and leads to more frequent failures during handover. PolicyConsensus~\cite{chen2025policyconsensus} and SafeDiff~\cite{wei2024ensuring}, in contrast, mainly focus on how actions should respond after a tactile change is detected. However, for chip handover, an abrupt tactile change typically indicates that the transfer is already in progress. At that point, reactive adjustments are often too late, because the policy has already produced an unsuitable handover orientation. TouchGuide instead aims to steer the handover actions to better satisfy contact physics constraints, and can more effectively leverage in-hand state to induce more appropriate handover angles and complete the task.

\subsubsection{Cucumber Peeling}
For the Cucumber Peeling task, the primary failure cases include large force and slippage, peeling hesitation, and peeling gap.

\textbf{Large Force and Slippage.} For bimanual cucumber peeling, effective coordination between the two arms is crucial. If the peeling arm applies excessive force, the other arm may lose its grip due to slippage and fail to hold the cucumber. Tactile sensing can provide reliable cues about whether excessive off-axis forces are applied and whether the gripper is slipping.

\textbf{Peeling Hesitation.} Without tactile information, the robot must rely on vision alone, making it difficult to determine when to initiate peeling or to sense whether the blade has engaged the cucumber. As a result, vision-only policies often exhibit hesitation at this stage.

\textbf{Peeling Gap.} Similarly, without tactile feedback, vision-only policies often maintain a gap from the cucumber surface when handling imperfectly smooth cucumbers, leading to poorer performance. In contrast, tactile sensing helps the policy recognize the current peeling contact condition and adjust accordingly.

As shown in Table~\ref{tab:results}, RDP achieves relatively high scores due to its high-frequency reactions to tactile changes. However, we also find that its failures often stem from an unfavorable contact angle when the peeler first touches the cucumber, which leads to partial peeling failures. In contrast, by incorporating contact-physics constraints, TouchGuide steers the policy toward more suitable actions, specifically a better initial peeling angle.

\subsubsection{Vase Wiping}
For the Vase Wiping task, the primary failure cases include incorrect grasp point, incorrect contact point, and large force.

\textbf{Incorrect Grasp Point.} This issue mainly arises when the initial position of the eraser changes. Without tactile feedback, the robot has difficulty perceiving the grasp location, which can lead to grasp failures. Moreover, it may not promptly detect an unsuccessful grasp or an incorrect grasp point during the attempt.

\textbf{Incorrect Contact Point.} This is largely due to missing in-hand state information. In the Vase Wiping task, without in-hand state the robot struggles to precisely adjust the wiping angle, often resulting in an incorrect point of force application and an improper wiping orientation. In-hand state can indicate the current grasp orientation of the eraser, enabling the robot to adjust its wiping strategy accordingly.

\textbf{Large Force.} Without tactile feedback, the robot may wipe with an improper angle and excessive force, which can cause the eraser to shift upward in the gripper. In that case, the gripper itself may directly contact the vase, leading to interference that prevents wiping regardless of the chosen wiping angle.

As a result, TouchGuide provides rich in-hand state and force information, completes the task with a higher success rate.

\subsubsection{Lock Opening}
For the Lock Opening task, common failure cases include incorrect grasp point, incorrect insertion point, and rotation with an insertion gap.

\textbf{Incorrect Grasp Point.} Similarly, without tactile sensing, it is difficult for the robot to accurately determine from the wrist camera whether the grasp is precise.

\textbf{Incorrect Insertion Point.} An incorrect insertion point is a key factor behind the low success rate on this task, and it primarily results from insufficient in-hand state information. In Lock Opening, the instant when the key first contacts the lock often largely determines whether the attempt will succeed. An incorrect contact angle can deflect the key or even knock it out of the gripper. In this case, changes in tactile feedback are not the most critical signal, because tactile input is typically sparse and nearly constant before contact. Instead, success hinges on the moment of initial contact. In-hand state can indicate where the key is grasped, enabling the policy to adjust the insertion angle and position accordingly.

\textbf{Rotation with Insertion Gap.} Another common failure case is initiating rotation before the key is fully inserted. From the top-down camera view, it is difficult to tell whether the key has been inserted all the way. In contrast, tactile feedback changes markedly when the key bottoms out and cannot be inserted further, which helps prevent premature rotation and improves the task success rate.

By providing rich in-hand state information and leveraging the strong generalization capability of $\pi_{0.5}$, TouchGuide built on $\pi_{0.5}$ can still achieve reasonably good success rates even in the low-data regime (\textbf{20 demonstration}) (see Tab.~\ref{tab:results}).

\subsubsection{Conclusion of Failure Cases}
\textbf{TouchGuide} provides the base policy with rich in-hand state information, thereby improving task success rates. Its CPM focuses on leveraging task-specific contact physics constraints to steer the policy toward actions that better satisfy contact physics. Compared to other visuo-tactile policies, TouchGuide has the advantage of producing more appropriate actions at the first contact, rather than only reacting after an abrupt tactile change. Moreover, thanks to its cross-policy nature, TouchGuide can be easily applied to tasks where $\pi_{0.5}$ performs well, requiring minimal modifications while achieving better generalization than diffusion-policy-based methods. As shown in Table~\ref{tab:results}, TouchGuide built on $\pi_{0.5}$~\cite{intelligence2025pi_} achieves a substantially higher success rate than TouchGuide built on Diffusion Policy~\cite{chi2025diffusion} (Force: \textbf{35.3\% $\rightarrow$ 49.9\%}, Tactile Image: \textbf{36.2\% $\rightarrow$ 58.0\%}).

\subsection{Additional Details on the Data Collection System}
\label{sec:appen:hardware}
In this section, we further provide details of the data collection system to complement our introduction in Sec.~\ref{sec:tacumi} and the experiments in Sec.~\ref{sec:exp:system}. Specifically, we compare our system with existing data collection systems, visualize action trajectories for SLAM-based UMI and our \textbf{TacUMI}, and describe TacUMI’s hardware design and component choices.

\begin{table}[!ht]
\vspace{-1mm}
\centering
\caption{A detailed comparison of existing data collection systems in terms of tactile feedback, precision, cost, and weight.}
\label{tab:hwcom}
\setlength{\tabcolsep}{6pt} 
\renewcommand{\arraystretch}{1.15}
\begin{tabular}{l l c c c c}
\toprule
& & & & Low-cost & Lightweight \\
Method & Category & Tactile Feedback &
\makecell{High-precision} &
\makecell{($<$ \$1000)} &
\makecell{($<$ 1000g)} \\
\midrule
GELLO~\cite{wu2024gello} & Teleop (Leader-follower) & None & \cmark & \cmark & \cmark \\
ALOHA~\cite{zhao2023act} & Teleop (Leader-follower) & None & \cmark & \xmark & \xmark \\
Bi-ACT~\cite{buamanee2024biact} & Teleop (Leader-follower) & Indirect (Force) & \cmark & \cmark & \cmark \\
Bunny-VisionPro~\cite{ding2025bunnyvisionpro} & Teleop (Hand Retargeting) & Indirect (Vibration) & \cmark & \cmark & \cmark \\
TactAR~\cite{xue2025reactive} & Teleop (VR Controller) & Indirect (Visual) & \cmark & \cmark & \cmark \\
UMI~\cite{chi2024umi} & Handheld (SLAM-based) & Semi-direct (Linkage) & \xmark & \cmark & \cmark \\
FastUMI~\cite{zhaxizhuoma2025fastumi} & Handheld (SLAM-based) & Semi-direct (Linkage) & \xmark & \cmark & \cmark \\
Touch in the Wild~\cite{zhu2025touch} & Handheld (SLAM-based) & Semi-direct (Linkage) & \xmark & \cmark & \cmark \\
ViTaMIn~\cite{liu2025vitamin} & Handheld (SLAM-based) & Semi-direct (Linkage) & \xmark & \cmark & \cmark \\
exUMI~\cite{xu2025exumi} & Handheld (VR Motion Tracker) & Semi-direct (Linkage) & \cmark & \cmark & \xmark \\
FARM~\cite{helmut2025farm} & Handheld (Motion Capture) & Semi-direct (Linkage) & \cmark & \xmark & \cmark \\
UMI-FT~\cite{choi2026umift} & Handheld (VIO ARKit) & Semi-direct (Linkage) & \xmark & \cmark & \cmark \\
ViTaMIn-B~\cite{li2025vitaminb} & Handheld (VR Motion Tracker) & \textbf{Direct (Rigid)} & \cmark & \cmark & \xmark \\
FreeTacMan~\cite{wu2025freetacman} & Handheld (Motion Capture) & \textbf{Direct (Rigid)} & \cmark & \xmark & \cmark \\
\hline
\textbf{TacUMI (Ours)} & Handheld (Vive Tracker) & \textbf{Direct (Rigid)} & \cmark & \cmark & \cmark \\
\bottomrule
\end{tabular}
\vspace{-3mm}
\end{table}

\begin{figure*}[!ht]
\vspace{-3mm}
  \centering
  \includegraphics[width=\textwidth]{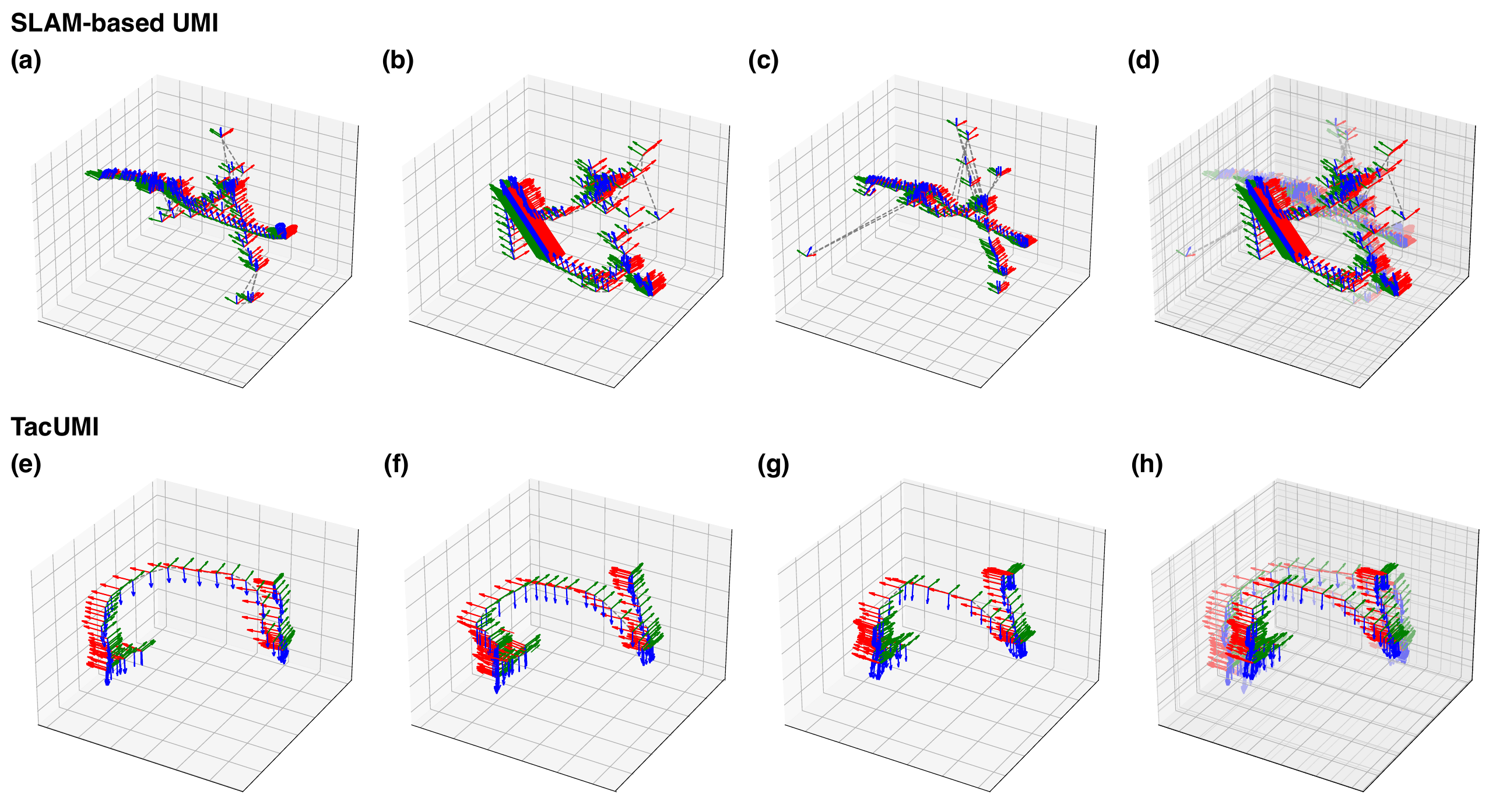}
  \caption{Trajectory visualization comparing SLAM-based UMI and \textbf{TacUMI}. (a)-(c) show three randomly sampled trajectories collected with SLAM-based UMI, and (e)-(g) show three randomly sampled trajectories collected with TacUMI. (d) and (h) overlay the trajectories from (a)-(c) and (e)-(g), respectively, illustrating trajectory consistency for the same task.}
  \label{fig:action_visualization}
  \vspace{-3mm}
\end{figure*}

\begin{figure*}[!ht]
  \centering
  \includegraphics[width=\textwidth]{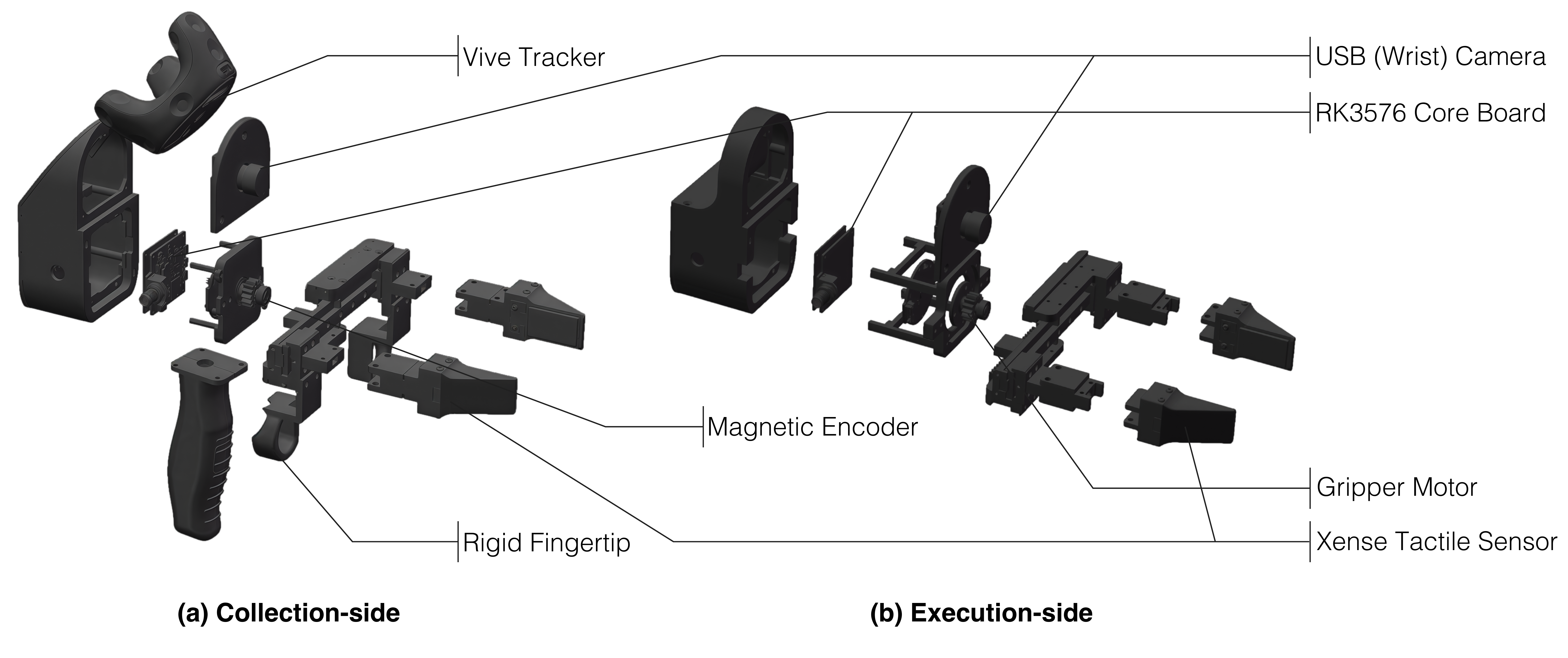}
  \caption{\textbf{TacUMI} Hardware Design. (a) Collection-side: TCP pose is directly provided; a Vive Tracker and a magnetic encoder measure end-effector pose and gripper position, respectively. (b) Execution-side: a gripper motor actuates the gripper, with an identical mechanical structure for direct deployment.
}
  \label{fig:hardware_design}
  \vspace{-8mm}
\end{figure*}

\subsubsection{Existing System Comparison}
\label{sec:appen:hardware:comparison}

As shown in Table~\ref{tab:hwcom}, we compare our system with state-of-the-art (SOTA) data collection systems. The evaluation criteria include \textbf{tactile feedback}, \textbf{precision}, \textbf{cost}, and \textbf{weight}.

\textbf{Tactile Feedback.} 
Regarding tactile feedback, some teleoperation systems either do not provide it~\cite{zhao2023act, wu2024gello} or only offer it indirectly~\cite{buamanee2024biact, ding2025bunnyvisionpro, xue2025reactive}. Specifically, Bi-ACT~\cite{buamanee2024biact} provides tactile feedback by applying force feedback to the master arm, Bunny-VisionPro~\cite{ding2025bunnyvisionpro} conveys contact signals through vibration, and TactAR~\cite{xue2025reactive} visualizes tactile information via the video stream in an VR headset. These forms of feedback are indirect, which is not well suited for collecting data that require fine-grained manipulation, such as handling fragile objects in Chip Handover or performing delicate tasks like Lock Opening and Shoe Lacing. In such settings, high-latency and indirect tactile feedback often leads to trajectories that exhibit hesitation or excessive force, ultimately degrading model performance. This limitation is clearly reflected in Table~\ref{tab:dc_succ}, where the policy success rates on these tasks highlight the shortcomings of teleoperation-based data collection. 

Moreover, most handheld grippers~\cite{chi2024umi,zhaxizhuoma2025fastumi,zhu2025touch,liu2025vitamin,xu2025exumi,helmut2025farm,choi2026umift} use a trigger to control gripper opening and closing. While this may be ergonomically convenient, it provides only semi-direct tactile feedback. With trigger mechanisms implemented via linkages and springs, it is difficult to directly perceive the force applied at the gripper fingertips, which often makes grasping fragile objects challenging.

In contrast, FreeTacMan~\cite{wu2025freetacman}, ViTaMIn-B~\cite{li2025vitaminb}, and our \textbf{TacUMI} adopt rigidly connected fingertip interfaces (see Fig.~\ref{fig:hardware_design}) to provide direct tactile feedback, which is crucial for fine-grained manipulation and for handling fragile or deformable objects.

\textbf{Precision.}
In terms of precision, SLAM-based UMI~\cite{chi2024umi,zhaxizhuoma2025fastumi,zhu2025touch,liu2025vitamin,choi2026umift} tends to exhibit lower precision, which can be insufficient for our tasks. In Fig.~\ref{fig:action_visualization}, we visualize the action trajectories collected by SLAM-based UMI and TacUMI, which we discuss in detail in Appendix~\ref{sec:appen:hardware:actiontrajvis}. Consistently, Table~\ref{tab:dc_succ} shows that policies trained on SLAM-based UMI data achieve substantially lower success rates than those trained on TacUMI data (\textbf{0\% $\rightarrow$ 20\%, 5\% $\rightarrow$ 30\%}).

Also, although VR Motion Tracker can achieve localization performance comparable to Vive Tracker, it requires the operator to wear a headset. As a result, tracking can be lost due to the operator’s movements or task-specific motions. In contrast, Vive Tracker relies on two external base stations for tracking and remains reliable for the vast majority of manipulation tasks. Moreover, because VR Motion Tracker requires operators to actively avoid tracking loss, they often act more cautiously during data collection, which increases the time per episode and leads to longer datasets. This advantage of Vive Tracker over VR Motion Tracker is evidenced in Table~\ref{tab:dc_user} by the number of attempts $\downarrow$ (\textbf{104 $\rightarrow$ 100}) required to collect 100 trajectories and the total dataset length $\downarrow$ (\textbf{14.7\,min $\rightarrow$ 13.5\,min}).

\textbf{Cost.}
In terms of cost, we consider a system (collection-side) priced below \$1,000 to be relatively low-cost. Our full system has a base cost of only \textbf{\$720}, which is much more affordable than solutions that rely on expensive motion capture (MoCap) devices~\cite{wu2025freetacman,helmut2025farm}. At the same time, it maintains good precision and usability, making it suitable for large-scale data collection without the environmental constraints imposed by MoCap systems.

\textbf{Weight.}
For weight, we consider the load borne by the operator. For example, systems that rely on VR~\cite{li2025vitaminb,xu2025exumi} require the user to wear a headset, and this weight is included in our evaluation to reflect the burden during prolonged use.
Our entire collection-side of \textbf{TacUMI} weighs only \textbf{540g}, making it comfortable for long-duration data collection. This is also one reason why TacUMI achieves higher user satisfaction $\uparrow$ than VR Motion Tracker in Table~\ref{tab:dc_user} (\textbf{9.0 $\rightarrow$ 9.6}).

\subsubsection{Action Trajectory Visualization}
\label{sec:appen:hardware:actiontrajvis}
Figure~\ref{fig:action_visualization} provides a detailed comparison between the action trajectories collected by SLAM-based UMI and TacUMI. In particular, Fig.~\ref{fig:action_visualization}(a)–(c) show three trajectories randomly sampled from the SLAM-based UMI data. We observe noticeable outliers, which arise because SLAM-based UMI can misestimate position and orientation when the end-effector is close to the tabletop or when the background provides few distinctive visual features. When feature points are severely lost, tracking may fail entirely. Although we can fill missing segments via linear interpolation, this inevitably discards information from the demonstration. For example, Fig.~\ref{fig:action_visualization}(b) corresponds to a case with substantial feature loss and extensive interpolation, making the underlying behavioral pattern difficult to discern from the resulting trajectory. Fig.~\ref{fig:action_visualization}(d) overlays the three trajectories in Fig.~\ref{fig:action_visualization}(a)–(c). We observe low consistency in the action trajectories collected with SLAM-based UMI, with frequent outliers and interpolation-induced loss of behavioral structure. Such artifacts are particularly detrimental for fine-grained manipulation.

In contrast, the action trajectories collected with \textbf{TacUMI} are visualized in Fig.~\ref{fig:action_visualization}(e)–(h). The first three plots show randomly sampled trajectories, and Fig.~\ref{fig:action_visualization}(h) overlays those in Fig.~\ref{fig:action_visualization}(e)–(g). We can see that TacUMI trajectories are smooth, with no visually apparent discontinuities or outliers, and they exhibit high consistency within the same task.

\subsubsection{\textbf{TacUMI} Hardware Design}
\label{sec:appen:hardware:design}

As shown in Fig.~\ref{fig:hardware_design}, TacUMI minimizes unnecessary mechanical structures to provide a high-quality data collection experience for the operator. As illustrated in Fig.~\ref{fig:hardware_design}(a), on the collection-side we adopt a Vive Tracker with two Lighthouse base stations, so the operator does not need to worry about tracking loss. To provide a unified software interface, we integrate an RK3576 core board on the gripper to process data in real time and stream high-resolution visual–tactile signals, and the mechanical design is highly integrated to reduce footprint as much as possible.

Moreover, fine-grained manipulation requires high precision in gripper position. For example, in Chip Handover, even a small error in the gripper opening can break the chip. We therefore avoid the ArUco-marker-based gripper opening estimation used by most UMI systems~\cite{chi2024umi,zhaxizhuoma2025fastumi,choi2026umift,li2025vitaminb,helmut2025farm,zhu2025touch,liu2025vitamin}, and instead employ a magnetic encoder to measure the gripper opening position. In addition, rigid fingertip interfaces provide direct tactile feedback, making it easier for the operator to perceive the current contact force.

Fig.~\ref{fig:hardware_design}(b) shows the execution-side of TacUMI, which must share the same mechanical structure as the collection-side to enable direct policy transfer.

\subsubsection{Hardware Selection Details}
As shown in Table~\ref{tab:hardware_components_quantity}, we provide the complete hardware bill of materials for TacUMI, covering all components required on both the collection-side and the execution-side, along with their quantities. Visual illustrations of selected components can be found in the assembly diagram in Fig.~\ref{fig:hardware_design}.

\begin{table}[!ht]
\vspace{-2mm}
\centering
\caption{\textbf{TacUMI} hardware bill of materials, including model numbers and quantities.}
\label{tab:hardware_components_quantity}
\begin{tabular}{cllc}
\toprule
 & Component & Model & Quantity \\
\midrule
\multirow{4}{*}{Collection-side}
& Tracker          & Vive Tracker        & 1 \\
& Base Station     & Vive Lighthouse     & 2 \\
& Magnetic Encoder & MT6835             & 1 \\
& Rigid Fingertip  & PLA 3D-Printed      & 2 \\
\midrule
Execution-side
& Gripper Motor    & GIM3505-8           & 1 \\
\midrule
\multirow{3}{*}{Both}
& Wrist Camera     & USB Camera          & 2 \\
& Onboard Compute Module & RK3576 Core Board & 2 \\
& Tactile Sensor         & Xense         & 4 / 2 (Shared) \\
\bottomrule
\end{tabular}
\vspace{-3mm}
\end{table}

Following careful hardware selection, \textbf{TacUMI} uses a wrist camera with a resolution of $640 \times 480$ and tactile images of $200 \times 350$. On the collection-side, the gripper tracking supports a maximum speed of $2\,\mathrm{m/s}$ and a maximum acceleration of $1\,\mathrm{m/s^2}$. On the execution-side, the gripper achieves a repeatability of $0.01\,\mathrm{mm}$, a maximum stroke of $85\,\mathrm{mm}$, a maximum opening and closing speed of $140\,\mathrm{mm/s}$, and a maximum gripping force of $40\,\mathrm{N}$. All observation and action data are synchronized via a unified software interface during collection, and are recorded at $30\,\mathrm{Hz}$.

This configuration enables \textbf{TacUMI} to collect high-precision tactile data. As shown in Fig.~\ref{fig:action_visualization}, the resulting trajectories are substantially cleaner and more consistent than those collected by other methods. As reported in Table~\ref{tab:dc_succ}, policies trained on TacUMI data achieve higher success rates compared to those trained on data collected with alternative systems, and TacUMI is also preferred by users (see Table~\ref{tab:dc_user}).

\subsection{\textbf{TouchGuide} Implementation Details}

\textcolor{touchblue}{Let Your Visuomotor Policies Have Touch Sensing!}

In this section, we provide a detailed implementation of TouchGuide in Alg.~\ref{alg:touchguide}. TouchGuide is designed to be largely decoupled from the base policy, which makes it easy to incorporate the tactile modality into a wide range of policies. We have described in detail in Sec.~\ref{sec:method:steering} how to apply TouchGuide to a policy, and Remark~\ref{remark:vlas} further notes that, when applied to VLAs, TouchGuide can operate on the action expert.

\subsubsection{\textbf{TouchGuide} Steers Diffusion Policy}
As shown in Code~\ref{code:dp}, steering DP with TouchGuide only requires adding four lines of \textcolor{touchblue}{blue} pseudocode at inference time, enabling fast and efficient tactile integration into a visuomotor policy with minimal overhead. Notably, we use an epsilon prediction diffusion policy, where the gradient is applied to the predicted noise. If one instead adopts a sample prediction formulation (\ie applying the gradient to the predicted sample rather than the noise), the coefficient in front of the gradient term must be re-derived and adjusted accordingly.

\begin{code}[!ht]
\vspace{-1mm}
\centering
\begin{Verbatim}[breaklines=true,fontsize=\small,
                 formatcom=\color{black},
                 codes={\catcode`\$=3\catcode`\^=7},
                 commandchars=\\\{\}]
\kw{# init}
x_t = randn_action_traj()
scheduler.set_timesteps(N)
\kw{# encode obs}
global_cond = obs_encoder(obs_seq).reshape(B, -1)
\kw{# diffusion sampling}
\kwd{for} step, t \kwd{in} enumerate(scheduler.timesteps):
    x_t[cond_mask] = cond_data[cond_mask]
    \kw{# predict noise}
    noise_pred = unet(x_t, t, global_cond)
    alpha = alphas_cumprod[t]
    \tb{# TouchGuide}
    \kwd{if} \tb{step < guidance_steps:}
        \tb{feasibility_score = contact_physical_model(x_t, obs_last)}
        \tb{g_t = grad(feasibility_score, x_t)}
        \tb{noise_pred = noise_pred - guidance_scale * sqrt(1 - alpha) * g_t}
    \kw{# scheduler step}
    x_t = scheduler.step(noise_pred, t, x_t).prev_sample
x_t[cond_mask] = cond_data[cond_mask]
\kwd{return} x_t
\end{Verbatim}
\caption{Illustrative pseudocode for \textbf{TouchGuide} steering of Diffusion Policy.}
\label{code:dp}
\vspace{-3mm}
\end{code}

\subsubsection{\textbf{TouchGuide} Steers $\pi_{0.5}$}
Similarly, for $\pi_{0.5}$, as shown in Code~\ref{code:pi05}, we incorporate tactile sensing by adding four lines of \textcolor{touchblue}{blue} pseudocode to the action expert at inference time. Note that our pseudocode follows the official implementation of $\pi_{0.5}$ and therefore uses the flow-matching direction $1 \rightarrow 0$. In practice, one may instead use the direction $0 \rightarrow 1$, in which case the coefficient in front of the gradient term must be re-derived. We refer readers to the derivation in Appendix~\ref{sec:appen:cg4fm} and omit further details here.

\begin{code}[!ht]
\vspace{-1mm}
\centering
\begin{Verbatim}[breaklines=true,fontsize=\small,
                 formatcom=\color{black},
                 codes={\catcode`\$=3\catcode`\^=7},
                 commandchars=\\\{\}]
\kw{# init}
x_t, t, dt = randn_actions(), 1.0, -1.0/N
\kw{# prefix -> KV cache}
img_tok = vision_tokens(obs.images)
txt_tok = text_tokens(obs.prompt)
prefix_tok = concat(img_tok, txt_tok)
prefix_mask = mask_prefix(obs)
kv_cache = build_kv_cache(prefix_tok, prefix_mask)
\kw{# action sampling process}
\kwd{for} step \kwd{in} range(N):
    \kw{# policy velocity v_t}
    a_tok = action_tokens(x_t)
    ada   = time_cond_pi05(t)
    h_tok = vlm_suffix(a_tok, kv_cache, prefix_mask, ada)
    v_t   = action_head(h_tok)
    \tb{# TouchGuide}
    \tb{\kwd{if} step < guidance_steps:}
    \kw{    # feasibility score}
    \tb{    feasibility_score = contact_physical_model(x_t, raw_obs)}
    \tb{    g_t = grad(feasibility_score, x_t)}
    \tb{    v_t = v_t - guidance_scale * g_t * (t/(1.0 - t))}
    x_t = x_t + dt * v_t
    t   = t + dt
\kwd{return} x_t
\end{Verbatim}
\caption{Illustrative pseudocode for \textbf{TouchGuide} steering of $\pi_{0.5}$.}
\label{code:pi05}
\vspace{-6mm}
\end{code}

\subsubsection{Performance of \textbf{TouchGuide}}
\label{sec:appen:tg_imple:performace}
As shown in Table 5, we compare the performance of $\pi_{0.5}$, DP, and their \textbf{TouchGuide} variants in terms of inference latency and mean task success rate. We use the JAX implementation of $\pi_{0.5}$, and all models follow a client–server setup, where the client handles data exchange with the robot and the server performs policy inference. We find that TouchGuide incurs only a small increase in inference latency while yielding a substantial improvement in task success rate.

\begin{table}[!ht]
\vspace{-1mm}
\centering
\caption{\textbf{TouchGuide} performance comparison on NVIDIA RTX PRO 6000 Blackwell.}
\label{tab:inference_time}
\begin{tabular}{lcccc}
\toprule
 & $\pi_{0.5}$ (JAX) & \textbf{TouchGuide ($\pi_{0.5}$)} & DP & \textbf{TouchGuide (DP)} \\
\midrule
Inference Speed $\uparrow$  & 18.52\,fps & 17.24\,fps (\textcolor{red}{$-$6.91\%}) & 12.82\,fps & 12.35\,fps (\textcolor{red}{$-$3.67\%}) \\
Average Success Rate $\uparrow$ & 35.9\% & 58.0\% (\textbf{$+$61.56\%}) & 16.3\% & 36.2\% (\textbf{$+$122.09\%}) \\
\bottomrule
\end{tabular}
\vspace{-3mm}
\end{table}

\subsection{Feasibility Score: Additional Physical Interpretation and Derivation}
\label{sec:appen:fs_derivation}
In this section, we provide a probabilistic motivation for the physical meaning of the feasibility score $\mathbf{s}$ introduced in Eq.~\ref{eq:feasibility_score_definition} in Sec.~\ref{sec:method:cpmtraining}. We also acknowledge DynaGuide~\cite{du2025dynaguide} for its probabilistic derivation of the distance metric, which inspired the design of our feasibility score. 

We follow DynaGuide and start from a very rough approximation that the latent space is Gaussian with diagonal covariance $\varSigma$, an assumption that is also implicitly adopted by methods based on squared latent space distances. Under this approximation, we model $p(\mathbf{a}_\text{real} | \mathbf{a}) = \mathcal{N} (\mathbf{a}_\text{real} | \mu=\mathbf{a},\varSigma=\sigma I)$, where $\mathbf{a_\text{real}}$ and $\mathbf{a}$ denote the latent representations of the optimal action sampled from the real distribution $\mathcal{Q}_\text{real}$ and an action $\mathbf{A}$ sampled from the policy distribution $\mathcal{Q}_\text{policy}$ (see Sec.~\ref{sec:method:steering}), respectively. The log probability can be expressed as $\log p(\mathbf{a}_\text{real} | \mathbf{a}) = - \frac{1}{2\sigma} \| \mathbf{a}_\text{real} - \mathbf{a} \|_2^2 + C$, where $C$ is a constant~\cite{du2025dynaguide}. Since the training objective of CPM is to align the latent observation $\mathbf{O}$ with $\mathbf{a}_\text{real}$ in the latent space, we can approximately obtain that 
\begin{align}
    \log p(\mathbf{a}_\text{real} | \mathbf{a}) \approx - \frac{1}{2\sigma} \| \mathbf{O} - \mathbf{a} \|_2^2 + C = - \frac{1}{2\sigma} \left( \| \mathbf{O} \|_2^2 + \| \mathbf{a} \|_2^2 - 2 \mathbf{O}^\top \mathbf{a} \right) + C.
\end{align}
Since the latent space is L2 normalized, the gradient of the log probability can be written as follows:
\begin{align}
    \nabla_\mathbf{A} \log p(\mathbf{a}_\text{real} | \mathbf{a}) = \frac{1}{\sigma} \nabla_\mathbf{A} \mathbf{O}^\top \mathbf{a} = \frac{1}{\sigma} \nabla_\mathbf{A} \mathbf{s}.
\end{align}
The constant prefactor $1/\sigma$ in front of the feasibility score $\mathbf{s}$ can be absorbed into the guidance scale $\eta$ we introduce. The above derivation provides a probabilistic motivation for the design of the feasibility score (see Eq.~\ref{eq:feasibility_score_definition} in Sec.~\ref{sec:method:cpmtraining}).

The above derivation of the feasibility score design provides a clear account of its physical meaning. The gradient of the feasibility score (see Eq.~\ref{eq:dp_steering} and Eq.~\ref{eq:fm_steering} in Sec.~\ref{sec:method:steering}) guides $\mathbf{a}$ toward $\mathbf{a}_\text{real}$, which explains why CPM encourages the base policy to produce actions that better comply with contact physics. It also substantiates our claim in Section~\ref{sec:method:steering} that CPM can steer the policy distribution $\mathcal{Q}_\text{policy}$ toward the real distribution $\mathcal{Q}_\text{real}$.

\subsection{Task-specific CPM Capability Analysis}
\label{sec:appen:inout_fea}

In this section, we visualize and analyze the feasibility scores produced by the task-specific Contact Physical Model (CPM). Implementation details of the CPM are provided in Sec.~\ref{sec:method:cpmtraining}, and the use of the feasibility score for steering the base policy is described in Sec.~\ref{sec:method:steering}.

\begin{figure*}[!ht]
\vspace{-1mm}
  \centering
  \includegraphics[width=0.8\textwidth]{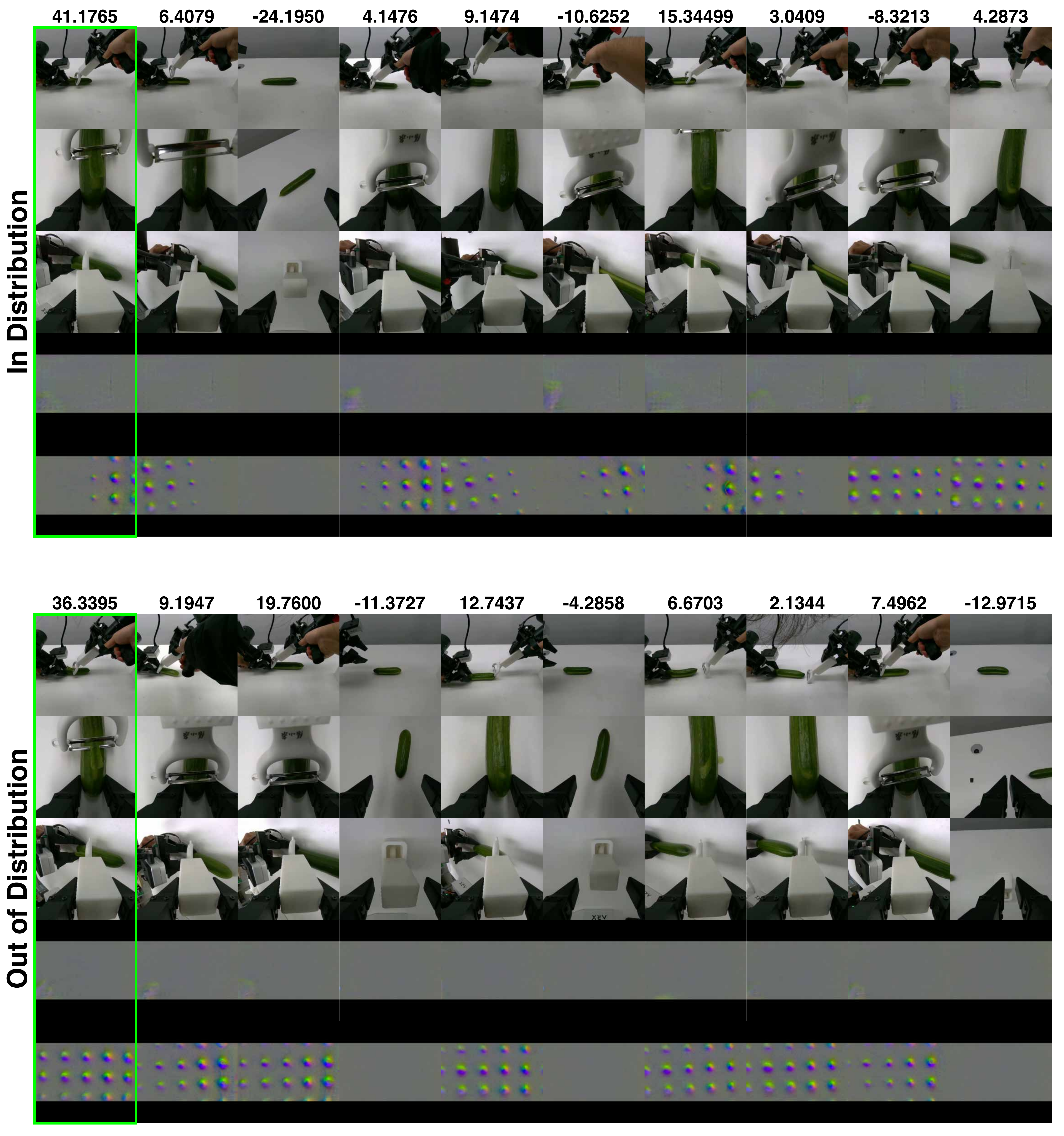}
  \caption{The CPM feasibility score under in-distribution and out-of-distribution settings. In-distribution (Top): The dataset is used for training. We report the feasibility score between actions in the \textcolor{gtgreen}{green} box and the outputs produced under observations randomly sampled from the dataset. Out-of-distribution (Bottom): Unseen object placements or orientations that the model was not trained on. The feasibility score is shown in the same way as above.}
  \label{fig:inoutfeasibility}
\end{figure*}

\begin{figure*}[!ht]
\vspace{-1mm}
  \centering
  \includegraphics[width=0.8\textwidth]{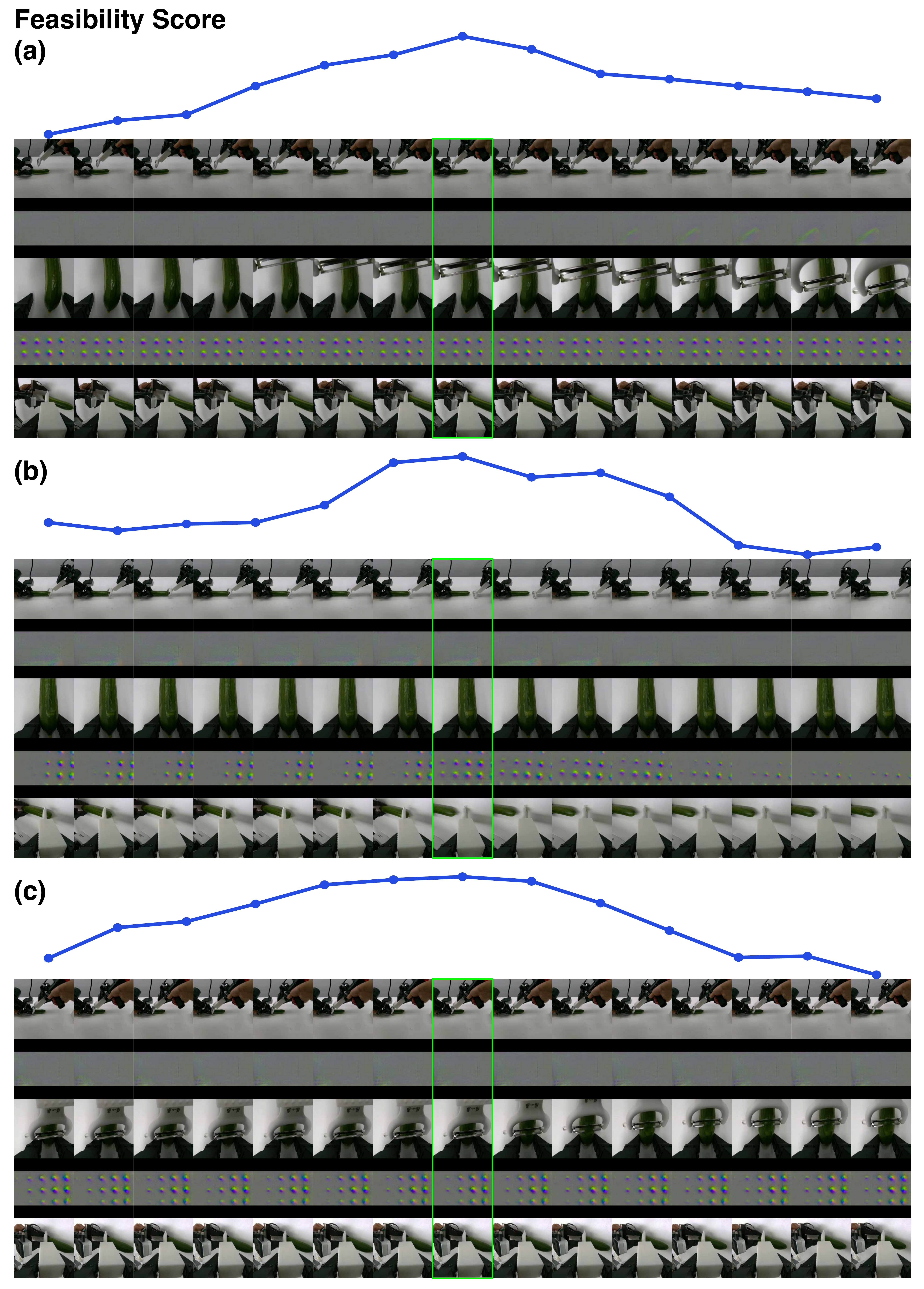}
  \caption{\textbf{Feasibility score visualization.} (a)–(c) We randomly sample from the dataset. The \textcolor{gtgreen}{green} box indicates the base frame we selected; we then expand a temporal window around it at 10 Hz, taking six frames before and six frames after, and visualize the corresponding feasibility scores (\textcolor{touchblue}{blue\,---}).}
  \vspace{-6mm}
  \label{fig:feasibility_validation}
\end{figure*}

\subsubsection{Out-of-Distribution Generalization}
As shown in Fig.~\ref{fig:inoutfeasibility}, we pair a randomly selected action with observations sampled at random and compute the resulting feasibility scores. From Fig.~\ref{fig:inoutfeasibility}(Top), we see that for in-distribution data, mismatched observations yield low feasibility scores, while the correctly paired observation receives a much higher feasibility score, which matches our expectation.

As shown in Fig.~\ref{fig:inoutfeasibility}, we also observe that the CPM produces clearly separable feasibility scores even for out-of-distribution (OOD) actions and observations with unseen placements or orientations. This suggests that our task-specific CPM has captured task-specific contact physics and exhibits a certain degree of OOD generalization.

\subsubsection{Task-specific Feasibility Score Validation}
We visualize the feasibility scores, as shown in Fig.~\ref{fig:feasibility_validation}(a)–(c).

Fig.~\ref{fig:feasibility_validation}(a) illustrates a case with sparse tactile variation. During peeling, because the human expert applies a roughly constant speed and force, the tactile images change only slightly, whereas the visual observations exhibit more pronounced changes. In contrast, Fig.~\ref{fig:feasibility_validation}(b) shows a visually sparse case near the end of Cucumber Peeling. As peeling gradually completes, only subtle differences are visible from the head camera, whereas the tactile images exhibit markedly different patterns.

Fig.~\ref{fig:feasibility_validation} matches our expectation in that the feasibility score gradually decreases as we move away from the base frame in both directions. This is because the collected actions are represented in the robot’s absolute joint space, where joint positions cannot change abruptly. Between two frames that are only $0.1\,s$ apart, the action typically changes only slightly, leading to a smooth decay in feasibility.

Despite the absence of abrupt action changes, we still observe clearly noticeable variations in the CPM outputs even for neighboring observations. This suggests that our CPM is sensitive enough for fine-grained manipulation, remaining effective even when action changes are small.

\subsection{Additional Experimental Analysis of the Action Distribution}
\label{sec:appen:tsne}
\begin{figure*}[!ht]
\vspace{-1mm}
  \centering
  \includegraphics[width=\textwidth]{figure/appen_tsne.pdf}
  \caption{Supplementary experiments on the Chip Handover (Hard) task. (a1) denotes the training data distribution of the expert demonstrations and the test setting for Chip Handover (Normal), while (a2) denotes the test setting for Chip Handover (Hard). (b) shows the performance of four representative baselines on the Chip Handover task under the normal and hard test settings. (c1)–(c4) present t-SNE visualizations of the action distributions. (c1) shows the three action distributions, namely $\mathbf{a}_\text{base}$, $\mathbf{a}_\text{steered}$, and $\mathbf{a}_\text{real}$. (c2) illustrates the distribution steering effect between $\mathbf{a}_\text{base}$ and $\mathbf{a}_\text{steered}$. (c3) and (c4) compare $\mathbf{a}_\text{base}$ with $\mathbf{a}_\text{real}$ and $\mathbf{a}_\text{steered}$ with $\mathbf{a}_\text{real}$, respectively.}
  \label{fig:appen_tsne}
  \vspace{-3mm}
\end{figure*}

To validate the generalization ability of the proposed method under out-of-distribution (OOD) settings, we introduce Chip Handover (Hard), as shown in Fig.~\ref{fig:appen_tsne}(a2), for evaluation. All baselines are trained on expert demonstrations collected under the setting in Fig.~\ref{fig:appen_tsne}(a1). As shown in Fig.~\ref{fig:appen_tsne}(b), \textbf{TouchGuide} achieves strong performance under both the normal and hard settings. Meanwhile, we observe that the use of TouchGuide makes the original baselines more robust to OOD scenarios, resulting in the smallest drop in success rate when transferring from the normal setting to the hard setting. This is consistent with our expectation. In Fig.~\ref{fig:inoutfeasibility} of Appendix~\ref{sec:appen:inout_fea}, we observe that CPM still achieves strong performance under OOD settings, indicating that it can improve the robustness of the policy when facing OOD scenarios.

Meanwhile, we conduct supplementary experiments to further verify the claim in Appendix~\ref{sec:appen:fs_derivation} that CPM can steer the base distribution $\mathcal{Q}_\text{base}$ toward the real distribution $\mathcal{Q}_\text{real}$ (see Sec.~\ref{sec:method:steering}). To obtain $\mathbf{a}_{\text{real}}$ from $\mathcal{Q}_\text{real}$, we follow the common assumption in imitation learning (IL) that expert demonstrations can be approximately regarded as the optimal actions $\mathbf{a}_{\text{real}}$~\cite{chi2025diffusion, hkummlab2025kai0}. We recollect data under the hard setting shown in Fig.~\ref{fig:appen_tsne}(a2) to obtain $\mathbf{a}_{\text{real}}$ (these data are not used for training any policy). We take $\pi_{0.5}$ as the base policy and denote the actions generated through its full sampling process as $\mathbf{a}_{\text{base}}$. Meanwhile, the actions produced with \textbf{TouchGuide} are denoted as $\mathbf{a}_{\text{steered}}$. We visualize these actions using t-SNE in Fig.~\ref{fig:appen_tsne}(c1)-(c4) to illustrate how CPM steers the action distribution. Fig.~\ref{fig:appen_tsne}(c2) shows the comparison between the distributions of $\mathbf{a}_{\text{base}}$ and $\mathbf{a}_{\text{steered}}$, from which we can observe that the steering effect of CPM on $\mathbf{a}_{\text{base}}$ is clearly reflected in the action distribution. Fig.~\ref{fig:appen_tsne}(c3) and (c4) present the distribution comparisons between $\mathbf{a}_{\text{base}}$ and $\mathbf{a}_{\text{real}}$, and between $\mathbf{a}_{\text{steered}}$ and $\mathbf{a}_{\text{real}}$, respectively. It can be clearly observed that, after being steered by CPM, the distribution of $\mathbf{a}_{\text{steered}}$ is closer to that of $\mathbf{a}_{\text{real}}$ than $\mathbf{a}_{\text{base}}$.

\subsection{Extended Implementation Details for Baselines}
\label{sec:appen:baselines}
To ensure a fair comparison, we apply the client–server setup to all baselines: the client is responsible only for exchanging data with the robot, while the server performs policy inference to improve inference throughput.

\textbf{Diffusion Policy.} We use the re-implementation of Diffusion Policy provided in RoboFactory (Official implementation: \href{https://github.com/real-stanford/diffusion_policy}{Diffusion Policy}~\cite{chi2025diffusion}, Re-implementation: \href{https://github.com/MARS-EAI/RoboFactory}{RoboFactory}~\cite{qin2025robofactory}).

\textbf{DP with Tactile Observation.} We use the same encoder as for visual observations to encode tactile observations, treating tactile input as an additional camera view and co-training the policy.

\textbf{SafeDiff.} In SafeDiff, force is used as the tactile signal. We re-implement the method by replacing the force input with tactile images and substituting the original tactile observation encoder accordingly.

\textbf{Reactive Diffusion Policy.} 
We use the official implementation of the tactile embedding (\href{https://github.com/xiaoxiaoxh/reactive_diffusion_policy}{Reactive Diffusion Policy}~\cite{xue2025reactive}).

\textbf{Policy Consensus.} We use the official implementation that includes a router model (\href{https://github.com/policyconsensus/policyconsensus}{Policy Consensus}~\cite{chen2025policyconsensus}).

\textbf{Tactile Dynamics \& DP.} The implementation of the Tactile Dynamics Model follows the dynamics model designs in Latent Policy Barrier and DynaGuide (\href{https://github.com/zhanyisun/lpb}{Latent Policy Barrier}~\cite{sun2025lpb}, \href{https://github.com/MaxDu17/DynaGuide}{DynaGuide}~\cite{du2025dynaguide}). Building on these, we augment the original dynamics model inputs by incorporating tactile observations. We then steer the DP using Eq.~\ref{eq:dp_steering}; detailed algorithmic descriptions are provided in Alg.~\ref{alg:touchguide}, and the pseudocode implementation is given in Code.~\ref{code:dp}.

\textbf{TouchGuide (Force) \& DP.}
The CPM implementation follows Sec.~\ref{sec:method:cpmtraining}. TouchGuide steers DP as shown in Sec.~\ref{sec:method:steering}, Alg.~\ref{alg:touchguide}, and Code.~\ref{code:dp}. The force version uses the software interface provided by the XenseSDK~\cite{xensesdk} to acquire tactile observations $\mathbf{T}^{\text{force}}_t \in \mathbb{R}^{35 \times 20 \times 3}$, where each element is a 3D force vector (\ie a 3D force vector field over a $35\times 20$ grid).

\textbf{TouchGuide (Tactile Image) \& DP.} The implementation is similar to the force version, except that the tactile image variant directly acquires tactile images $\mathbf{T}_t^\text{image} \in \mathbb{R}^{350 \times 200 \times 3}$ from the XenseSDK~\cite{xensesdk}.

\textbf{$\pi_{0.5}$.} We use the official implementation without any modifications (\href{https://github.com/Physical-Intelligence/openpi}{$\pi_{0.5}$}~\cite{intelligence2025pi_}).

\textbf{$\pi_{0.5}$ with Tactile Observation.} To use the official pretrained weights consistently, we directly concatenate the tactile images and wrist camera images, preserving tactile information without modifying the model architecture.

\textbf{Tactile Dynamics \& $\pi_{0.5}$.}
Similar to the Tactile Dynamics setup above, we perform guidance using Eq.~\ref{eq:fm_steering}; details are provided in Alg.~\ref{alg:touchguide} and Code.~\ref{code:pi05}.

\textbf{TouchGuide (Force) \& $\pi_{0.5}$.}
Similar to the implementation above, the details are provided in Alg.~\ref{alg:touchguide} and Code.~\ref{code:pi05}.

\textbf{TouchGuide (Tactile Image) \& $\pi_{0.5}$.} Similar to the force version above.

\subsection{Extended Task Descriptions}
\label{sec:appen:tasks}
In this section, as a supplement to Sec.~\ref{sec:exp:setup}, we provide a detailed introduction to five challenging fine-grained manipulation tasks, including more comprehensive task descriptions, their key challenges, and the target capabilities of the tactile sensor.

\subsubsection{Task Challenges and Targeted Capabilities}
Below we provide additional details for each of the five tasks.
\label{sec:appen:tasks:detail}

\textbf{Shoe Lacing.} A 3D-printed fixture is used to hold the shoe in place. The shoelace is untied and placed on the 3D-printed lace clamp. The initial lace pose varies: it may be positioned slightly forward or backward on the clamp, and its pitch angle can vary within a limited range (approximately $\pm 30^\circ$).

First, the left arm grasps the shoelace and threads it through the eyelet. This step requires the policy to effectively leverage the in-hand state: since the initial lace pose varies, the grasped position may differ slightly across trials, and the policy must adjust the threading direction accordingly. After insertion, the left arm continuously adjusts its motion to feed more of the shoelace through the eyelet. Once a sufficient length has been fed through, the right arm pulls the lace out. At this point, the lace orientation is often unsuitable for threading through the next eyelet, so the two arms must repeatedly adjust until the lace is properly positioned in the right gripper. This step requires tactile feedback: only the lace tip is stiff enough for threading. Grasping too close to the tip leaves insufficient length for insertion, while grasping too far back means holding a softer segment that cannot pass through the eyelet. After the adjustment, the system proceeds to the next eyelet. The left and right arms then alternate to repeat the above procedure four times, after which the task terminates.

\textbf{Chip Handover.}
Initially, a potato chip is placed on the chip-can lid and randomly positioned within the robot’s reachable workspace, while the chip can remains largely fixed. The right arm first grasps the chip from the table. During grasping, tactile feedback provides force information to prevent the chip from being crushed.

Next, the right arm moves to a suitable handover pose in midair, and the two arms coordinate to transfer the chip. The left arm grasps the chip near the edge at a relatively flat region. This step requires tactile feedback to provide the in-hand state: even slight grasping deviations can cause the handover to fail, so the precision requirement is extremely high.

Finally, the left arm places the chip into the chip can. The task is considered complete if the chip remains unbroken, is successfully handed over once, and is ultimately placed inside the can.

\begin{figure*}[!ht]
\vspace{-3mm}
  \centering
  \includegraphics[width=\textwidth]{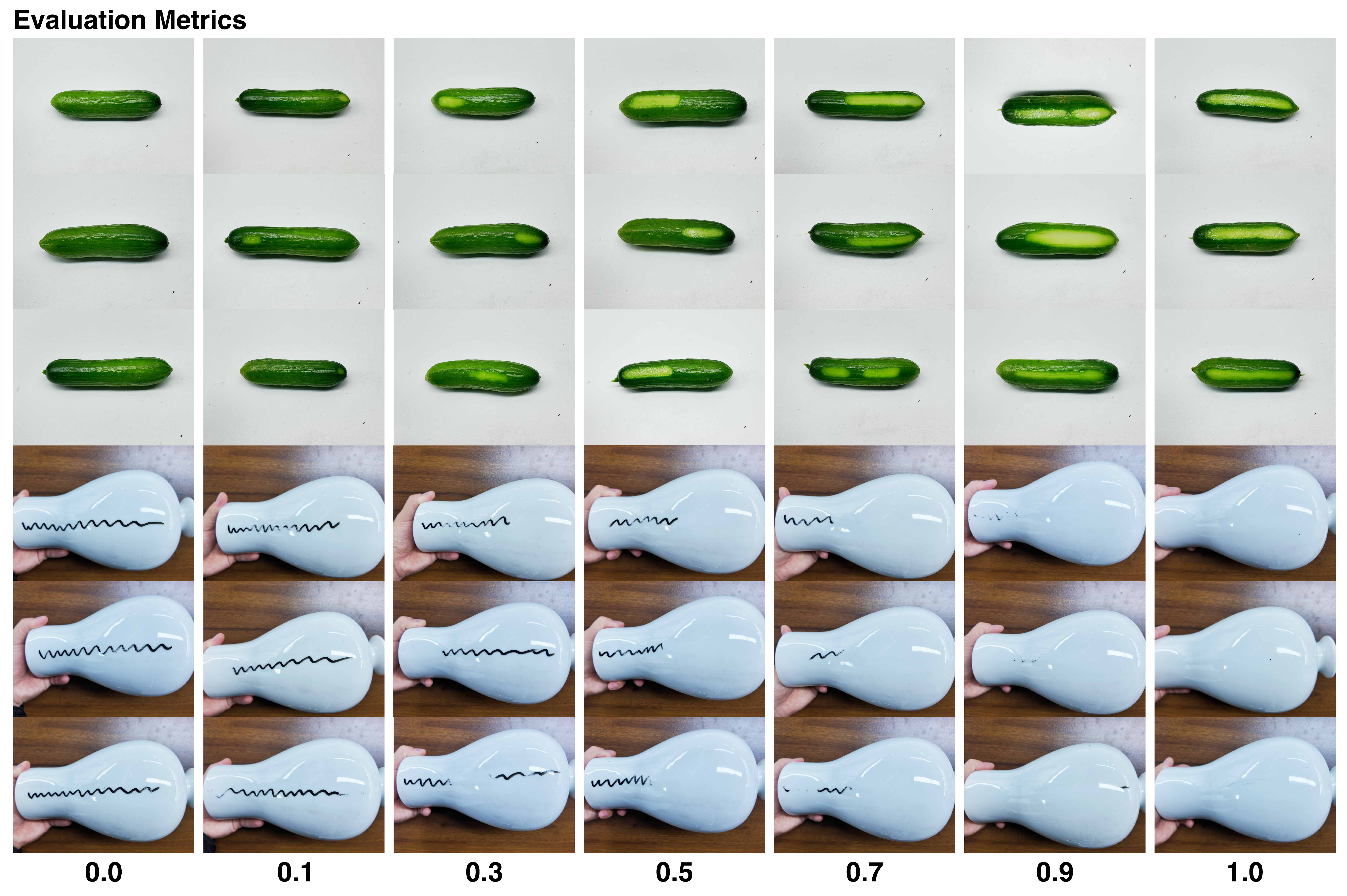}
  \caption{Evaluation metrics for the Cucumber Peeling and Vase Wiping tasks.}
  \label{fig:evaluationmetric}
  \vspace{-3mm}
\end{figure*}

\textbf{Cucumber Peeling.} 
Initially, the peeler is largely fixed, while the cucumber is randomly placed within the robot’s reachable workspace with a randomized yaw angle (approximately $\pm 30^\circ$).

The right arm first picks up the peeler. Both arms then move toward the cucumber. After the left arm firmly grasps the cucumber, the right arm brings the peeler close and embeds the blade head into the cucumber. This step requires tactile feedback to determine whether the gripper is firmly holding the cucumber without damaging it. Tactile feedback is also necessary for inserting the blade head. Without tactile sensing, policies typically hesitate at this stage.

The two arms then collaborate to peel continuously until an entire strip of cucumber skin is removed intact. This stage requires ongoing tactile feedback. For the left arm, tactile sensing is needed to detect whether the cucumber is slipping and to maintain a secure grasp without damaging it. For the right arm, tactile feedback helps the peeler maintain an appropriate angle and apply smooth, even peeling motion. The task terminates once a full strip of cucumber peel has been removed.

\textbf{Vase Wiping.}
Initially, the vase is held by a human and randomly rotated by a small angle (approximately $15^\circ$). The eraser is randomly placed within the workspace. The robot must first grasp the eraser. This step requires tactile feedback to verify a firm grasp and a correct contact location: if the robot grasps the soft lower part of the eraser, it will fail to erase the markings during wiping.

The robot then wipes along the vase’s curved surface with the eraser to remove the markings. This step requires tactile feedback to conform to the vase surface while avoiding excessive force. If too much force is applied, the eraser can be knocked out of alignment, preventing subsequent wiping. The task terminates once the vase has been completely wiped clean.

\textbf{Lock Opening.}
Initially, the key and the lock are held by a human and randomly positioned within the robot’s workspace. The robot must first grasp the key. This step requires tactile feedback to ensure the key is securely held.

Next, the robot moves toward the lock and inserts the key. This step requires rich tactile in-hand state feedback. Because the grasp location on the key varies across trials, the robot must adjust the insertion angle accordingly. If the initial contact is significantly misaligned, the key may slip or be knocked off-axis, substantially reducing the success rate of subsequent attempts.

After the key is fully inserted, the robot rotates it to unlock. This step requires the robot to sense whether the key has been inserted all the way. If it starts turning before full insertion, the gripper or the lock may be damaged. The task terminates once the lock is successfully opened.

\subsubsection{Evaluation Metrics Details}
For Shoe Lacing, Chip Handover, and Lock Opening, we evaluate performance using the success rate. The task termination conditions have been described in Appendix~\ref{sec:appen:tasks:detail}. For each trial, we set a maximum time limit for task completion, typically three times the average episode length of human expert demonstrations. If the task is not completed within this time limit, the trial is considered a failure.

For Cucumber Peeling and Vase Wiping, we evaluate performance using a task score. As shown in Fig.~\ref{fig:evaluationmetric}, the score is defined as the robot’s completion divided by the completion achieved by a human expert. For Cucumber Peeling, the score is the ratio of the peeled cucumber-skin length. For Vase Wiping, it is the ratio of the erased marking length.

\subsection{Training Hyperparameters}
\subsubsection{Base Policy}
DP training hyperparameters are shown in Table~\ref{tab:dp_hparas}, and $\pi_{0.5}$ training hyperparameters are shown in Table~\ref{tab:pi_hparas}.

\begin{table}[ht]
\vspace{-3mm}
\centering
\small
\caption{Hyperparameters for DP training on NVIDIA RTX PRO 6000 Blackwell.}
\setlength{\tabcolsep}{3pt}
\renewcommand{\arraystretch}{1.15}
\begin{tabular}{cccccc}
\toprule
 \textbf{Hyperparameters} & \textbf{Shoe Lacing} & \textbf{Chip Handover} & \textbf{Cucumber Peeling} & \textbf{Vase Wiping} & \textbf{Lock Opening} \\
\midrule
\makecell{Image Shape \\ (Resolution $\times$ Views)} & (224, 224, 3) $\times$ 3 & (224, 224, 3) $\times$ 3 & (224, 224, 3) $\times$ 3 & (224, 224, 3) $\times$ 1 & (224, 224, 3) $\times$ 1 \\
Action Shape & 14 & 14 & 14 & 10 & 10\\
Batch Size & 32 & 32 & 32 & 32 & 32 \\
Learning Rate & 1.0e-4 & 1.0e-4 & 1.0e-4 & 1.0e-4 & 1.0e-4 \\
Warm-up Steps & 500 & 500 & 500 & 500 & 500 \\
Betas & [0.95, 0.999] & [0.95, 0.999] & [0.95, 0.999] & [0.95, 0.999] & [0.95, 0.999] \\
Weight Decay & 1.0e-6 & 1.0e-6 & 1.0e-6 & 1.0e-6 & 1.0e-6 \\
Epsilon & 1.0e-8 & 1.0e-8 & 1.0e-8 & 1.0e-8 & 1.0e-8 \\
Epochs & 300 & 300 & 300 & 300 & 300 \\
\bottomrule
\end{tabular}
\label{tab:dp_hparas}
\vspace{-3mm}
\end{table}

\begin{table}[ht]
\vspace{-3mm}
\centering
\small
\caption{Hyperparameters for $\pi_{0.5}$ training NVIDIA RTX PRO 6000 Blackwell.}
\setlength{\tabcolsep}{3pt}
\renewcommand{\arraystretch}{1.15}
\begin{tabular}{cccccc}
\toprule
 \textbf{Hyperparameters} & \textbf{Shoe Lacing} & \textbf{Chip Handover} & \textbf{Cucumber Peeling} & \textbf{Vase Wiping} & \textbf{Lock Opening} \\
\midrule
\makecell{Image Shape \\ (Resolution $\times$ Views)} & (224, 224, 3) $\times$ 3 & (224, 224, 3) $\times$ 3 & (224, 224, 3) $\times$ 3 & (224, 224, 3) $\times$ 1 & (224, 224, 3) $\times$ 1 \\
Action Shape & 14 & 14 & 14 & 10 & 10\\
Batch Size & 64 & 64 & 64 & 64 & 64 \\
Paligemma & \texttt{gemma\_2b\_lora} & \texttt{gemma\_2b\_lora} &\texttt{gemma\_2b\_lora}& \texttt{gemma\_2b\_lora} & \texttt{gemma\_2b\_lora} \\
Action Expert & \texttt{gemma\_300m\_lora}& \texttt{gemma\_300m\_lora}& \texttt{gemma\_300m\_lora}& \texttt{gemma\_300m\_lora}& \texttt{gemma\_300m\_lora}\\
Weight & \texttt{pi05\_base} & \texttt{pi05\_base}& \texttt{pi05\_base} & \texttt{pi05\_base} & \texttt{pi05\_base} \\
EMA Decay & 0.99 & 0.99 & 0.99 & 0.99 & 0.99 \\
Steps & 40,000 & 20,000 & 20,000 & 20,000 & 20,000 \\
\bottomrule
\end{tabular}
\vspace{-3mm}
\label{tab:pi_hparas}
\end{table}

\subsubsection{Task-specific CPM}
CPM training hyperparameters are shown in Table~\ref{tab:cpm_hparas}.
\begin{table}[ht]
\vspace{-3mm}
\centering
\small
\caption{Hyperparameters for CPM training NVIDIA RTX PRO 6000 Blackwell.}
\setlength{\tabcolsep}{3pt}
\renewcommand{\arraystretch}{1.15}
\begin{tabular}{cccccc}
\toprule
 \textbf{Hyperparameters} & \textbf{Shoe Lacing} & \textbf{Chip Handover} & \textbf{Cucumber Peeling} & \textbf{Vase Wiping} & \textbf{Lock Opening} \\
\midrule
\makecell{Image Shape \\ (Resolution $\times$ Views)} & (224, 224, 3) $\times$ 3 & (224, 224, 3) $\times$ 3 & (224, 224, 3) $\times$ 3 & (224, 224, 3) $\times$ 1 & (224, 224, 3) $\times$ 1 \\
\makecell{Tactile Shape \\ (Resolution $\times$ Views)} & (224, 224, 3) $\times$ 2 & (224, 224, 3) $\times$ 2 & (224, 224, 3) $\times$ 2 & (224, 224, 3) $\times$ 2 & (224, 224, 3) $\times$ 2 \\
Action Shape & 14 & 14 & 14 & 10 & 10\\
Batch Size & 64 & 64 & 64 & 64 & 64 \\
Learning Rate & 1.0e-5 & 1.0e-5 & 1.0e-5 & 1.0e-5 & 1.0e-5 \\
Warm-up Ratio & 0.05 & 0.05 & 0.05 & 0.05 & 0.05 \\
Epochs & 200 & 200 & 200 & 200 & 200 \\
\bottomrule
\end{tabular}
\label{tab:cpm_hparas}
\vspace{-3mm}
\end{table}

\end{document}